\documentclass[journal]{IEEEtran}
\IEEEoverridecommandlockouts
\usepackage{times}
\usepackage{amsmath}
\usepackage{amssymb}
\usepackage{physics}
\usepackage{cite}
\usepackage{cleveref}
\usepackage{tabularx}
\usepackage{multirow}
\usepackage{algorithm}
\usepackage{algpseudocode}
\usepackage{booktabs}
\usepackage{xcolor}
\usepackage{makecell}
\usepackage{fancyhdr}
\usepackage{graphicx}
\usepackage{subcaption}
\usepackage[font=footnotesize]{caption}

\DeclareMathAlphabet\mathzapf {T1}{pzc} {mb} {it}
\usepackage{array}
\newcolumntype{L}[1]{>{\raggedright\let\newline\\\arraybackslash\hspace{0pt}}m{#1}}
\newcolumntype{C}[1]{>{\centering\let\newline\\\arraybackslash\hspace{0pt}}m{#1}}
\newcolumntype{R}[1]{>{\raggedleft\let\newline\\\arraybackslash\hspace{0pt}}m{#1}}

\DeclareGraphicsExtensions{.pdf,.jpeg,.png}

\Crefname{figure}{Fig.}{Figs.}

\title{MarUco: A Markerless 6D Pose Estimation Framework for Closed-Loop Control of Surgical Continuum Manipulators}

\author{Junhyun Park$^{*}$, Chunggil An$^{*}$, Myeongbo Park, Ihsan Ullah, Sihyeong Park, and Minho Hwang$^{\dagger}$%
\thanks{$^{*}$These authors contributed equally. $^{\dagger}$Corresponding author.}%
\thanks{C. An, M. Park, I. Ullah, S. Park, and M. Hwang are with the Department of Robotics and Mechatronics Engineering, DGIST, Daegu 42988, Republic of Korea (e-mail: \{cndrlfwlq, qkraudqh23, ihsankhan, psh120, minho\}@dgist.ac.kr).}%
\thanks{J. Park conducted this work with the Department of Robotics and Mechatronics Engineering, DGIST, Daegu 42988, Republic of Korea, and is currently affiliated with the AI Research Lab, DEEPNOID, Seoul, Republic of Korea (e-mail: junhyunpark0507@gmail.com).}}

\date{}

\begin{document}
\maketitle

%%%%%%%%%%%%%%%%%%%%%%%%%%%%%%%%%%%%%%%%%%%%%%%%%%%%%%%%%%%%%%%%%%%%%%%%%%%%%%%%
\begin{abstract}
Flexible endoscopic continuum manipulators offer high dexterity and access to
complex anatomy, but nonlinear hysteresis limits feedforward control accuracy.
Closed-loop control can compensate for these errors but requires accurate
six-degree-of-freedom (6D) end-effector pose feedback. We present MarUco, a
markerless 6D pose estimation framework for closed-loop control using only
stereo vision during operation. A photorealistic pseudo-rigid-body simulation
pipeline generates large-scale annotated training data without manual labeling.
A multifeature fusion network integrates masks, keypoints, heatmaps, and
bounding boxes from both stereo views to estimate an initial pose, followed by
a learned single-pass render-and-compare module that refines the pose without
iterative optimization. Kinematics-free hand--eye calibration estimates
camera-to-robot-base extrinsics, and self-supervised adaptation uses unlabeled
real stereo pairs to mitigate sim-to-real pose error. Across 1{,}000 real
samples, MarUco achieves translation and rotation errors of
$0.78 \pm 0.50$\,mm and $3.07 \pm 1.25^\circ$, respectively, with a total
processing time of 52.4\,ms per stereo pair. In closed-loop experiments over
eight reference paths, MarUco achieves a mean terminal translation error of
1.8\,mm, an 88\% reduction relative to uncompensated open-loop control. To the
best of our knowledge, this is the first markerless position-based visual
servoing framework for continuum manipulators.
\end{abstract}

%%%%%%%%%%%%%%%%%%%%%%%%%%%%%%%%%%%%%%%%%%%%%%%%%%%%%%%%%%%%%%%%%%%%%%%%%%%%%%%%


\section{Introduction}
 Flexible endoscopic surgical systems employ cable-driven continuum manipulators whose compliant backbones navigate curvilinear paths and access complex anatomy through natural-orifice routes, including gastrointestinal and transoral approaches~\cite{Hwang2020KFLEXAF, Remacle2015TransoralRS, Phee2009MasterAS, Zorn2018ANT, park2024sam}. These manipulators are typically driven by meter-scale tendon--sheath transmissions connected to remotely located motors. Friction, backlash, cable elongation, and twisting within these transmissions introduce hysteresis~\cite{Roy2017ModelingAE,Kim2022RecurrentNN,Ji2020AnalysisOT,Dalvand2018AnAL,park2024hysteresis}, causing a discrepancy between the commanded joint configuration and the actual end-effector pose.

Feedforward compensation addresses this discrepancy by adjusting the joint commands based on a model of hysteresis. Analytical approaches explicitly model the underlying physical effects. However, accurately modeling the multiple nonlinear effects that constitute hysteresis is challenging, and the resulting models are often specific to a particular hardware configuration, limiting their generalizability across different systems~\cite{Kim2021ShapeadaptiveHC, Lee2020NonLinearHC}. Learning-based approaches can capture multiple nonlinear effects jointly and can be adapted to new hardware through retraining~\cite{park2024sam, park2024hysteresis}. However, because feedforward compensation does not measure the actual end-effector pose, residual errors cannot be directly observed and corrected.

Closed-loop control instead uses feedback of the actual end-effector state to correct these residual errors~\cite{chaumette2006visual}. Its accuracy therefore depends critically on the quality of end-effector pose feedback. For continuum manipulators, whose end-effector pose cannot be reliably inferred from actuation-side measurements because of hysteresis, several sensing approaches have been explored.

Marker-based vision can provide direct pose measurements~\cite{wu2021closedloop}. However, visual markers are susceptible to occlusion and contamination by blood in surgical environments~\cite{zhang2017realtime, tsui2023optical}, while rigid fiducial markers are difficult to integrate with millimeter-scale instruments and introduce mechanical deflection due to their added mass~\cite{park2024sam}.

Embedded sensing provides an alternative without external visual markers. Fiber Bragg grating (FBG) sensors reconstruct instrument shape from strain measured along embedded optical fibers, but require specialized interrogation hardware and careful mechanical integration~\cite{7723875, FBG_deflection, khan2020pose}. Electromagnetic (EM) tracking directly localizes sensor coils at the instrument tip, but its accuracy can degrade in the presence of metallic objects or electrosurgical interference~\cite{franz2014emtracking, sun2023eminterference}.

In contrast, markerless vision estimates the end-effector pose directly from camera images without modifying the instrument or attaching markers. Cameras are already integrated into flexible endoscopic platforms~\cite{Hwang2020KFLEXAF, Remacle2015TransoralRS, Phee2009MasterAS, Zorn2018ANT}, including stereo endoscopes used for depth perception in systems such as the da Vinci~\cite{dvrk}. Markerless vision therefore provides a practical route to end-effector pose feedback for closed-loop control.

In this work, we estimate the full 6D pose rather than individual 3D image features because the control objective is defined by both the position and orientation of the tool center point (TCP). Moreover, the TCP is a virtual task frame determined by the dual-jaw geometry rather than a single directly observable point. Representing the feedback as a pose in $SE(3)$ therefore allows the desired task-space configuration to be regulated directly using position-based visual servoing (PBVS).

 However, achieving markerless closed-loop control in practice requires addressing three challenges: scalable generation of training data, fast and accurate pose estimation, and reliable extrinsic calibration between the vision system and the robot.

First, training a markerless pose estimator requires large-scale 6D pose annotations, which are difficult to obtain for continuum manipulators. Prior approaches have relied on manual annotation or heuristic image-based labeling of real images~\cite{reiter2014learning, spektor2024monocular, wang2023}, while synthetic generation for continuum instruments has primarily used custom domain-randomized rendering for monocular pose estimation~\cite{zhou2024markerless}. Scalable generation of photorealistic stereo data with rich geometric annotations therefore remains underexplored.

Second, pose estimation must be both accurate and sufficiently fast for closed-loop feedback. Feedforward networks provide low-latency inference but do not explicitly enforce geometric consistency between the estimated pose and the observed image~\cite{Ye_2016}. Render-and-compare methods improve geometric consistency by aligning rendered silhouettes or keypoints with observed features, but conventional approaches require iterative optimization, increasing inference latency. Zhou \textit{et al.} reported a translation error of 1.24\,mm after iterative refinement, with the refinement stage alone requiring approximately 800\,ms per frame~\cite{zhou2024markerless}.

Third, deployment requires accurate extrinsic calibration between the cameras and the robot base. Conventional hand--eye calibration typically relies on end-effector poses obtained from robot kinematics. For tendon-driven continuum manipulators, however, hysteresis makes these kinematic poses unreliable, limiting the applicability of standard kinematics-based calibration. Uncertainty in the estimated camera-to-base transformation can also contribute to systematic errors in robot-frame pose estimation and consequently degrade closed-loop control accuracy.

To address these challenges, we propose MarUco (\textbf{Mar}kerless \textbf{U}nified pose estimation framework for the closed-loop control of \textbf{co}ntinuum manipulators), and validate the complete framework on real hardware.

\begin{itemize}

\item \textbf{Automatic photorealistic data generation.}
We develop a simulation pipeline that represents the continuum manipulator using a pseudo-rigid-body URDF and generates 200{,}000 annotated stereo pairs in Isaac Sim. The pipeline provides masks, keypoints, bounding boxes, and 6D pose labels under domain randomization without manual annotation.

\item \textbf{Low-latency pose estimation with single-pass render-and-compare refinement.}
We introduce a multifeature fusion network (MFFN) that integrates geometric features from both stereo views to estimate an initial 6D pose. A subsequent feedforward refinement module compares the predicted geometric features with those rendered at the initial pose and directly regresses a residual pose correction in a single pass, avoiding iterative optimization. The resulting estimator achieves $0.78$\,mm translation error and $3.07^\circ$ rotation error at 52.4\,ms per stereo pair.

\item \textbf{Kinematics-free hand--eye calibration and self-supervised real-domain adaptation.}
We estimate the camera-to-robot-base extrinsics without relying on continuum manipulator forward kinematics by combining RGB-D-based fiducial localization with ChArUco-based pose measurements from the stereo cameras. We further mitigate residual systematic pose error through self-supervised adaptation using 150 unlabeled real stereo pairs without manual pose annotation.

\item \textbf{Markerless position-based visual servoing.}
We integrate the proposed pose estimator with position-based visual servoing and validate the complete framework on a physical continuum manipulator. Across eight reference paths, MarUco achieves a mean terminal translation error of 1.8\,mm, corresponding to an 88\% reduction relative to uncompensated open-loop control. To the best of our knowledge, this is the first markerless position-based visual servoing framework for continuum manipulators.

\end{itemize}

\section{Related Works}
 We review prior work along the four axes most relevant to MarUco: closed-loop visual control, pose estimation of continuum manipulators, synthetic data generation, and rendering-based pose refinement.

\subsection{Closed-Loop Control of Continuum Manipulators}
Closed-loop control of continuum manipulators has been investigated using visual servoing~\cite{CASE_vision_continuum_servoing, nazari2022visual_servo_continuum, chen2024stereo_visual_servo, ivs_goldberg}, dynamics-based control~\cite{campisano2021closedloop, wang2021dynamic, doroudchi2022configuration}, and reinforcement learning~\cite{ji2021safe, 8531756}. These approaches require feedback of the manipulator state.

Prior markerless vision-based control of continuum and concentric-tube robots has primarily relied on image-based visual servoing (IBVS), which regulates errors directly in image-feature space without explicitly recovering the end-effector pose~\cite{CASE_vision_continuum_servoing, nazari2022visual_servo_continuum}. IBVS has the advantage of reduced sensitivity to calibration errors because the control error is defined directly in image space. However, image-space control can be affected by image-Jacobian singularities and local minima~\cite{chaumette1998problems, chaumette2006visual, mppi_vs2021, stereo_ibvs_localminima2025}, and task-space pose targets must be represented indirectly through desired image features~\cite{azizian2014survey}.

PBVS instead regulates an explicit task-space pose error in $SE(3)$, but requires reliable 6D pose feedback~\cite{chaumette2006visual}. Prior closed-loop control systems requiring end-effector pose feedback have obtained it from dedicated sensors or external markers~\cite{penning2012closedloop, sefati2021fbg, wu2021closedloop}. MarUco addresses this gap by providing markerless 6D pose feedback for position-based visual servoing of a continuum manipulator.

\subsection{Pose Estimation of Continuum Manipulators}
End-effector pose estimation for continuum manipulators has been approached through embedded sensing, marker-based vision, and markerless vision.

Embedded sensing approaches estimate the distal state using FBG sensors, electromagnetic trackers, or proximal self-sensing~\cite{khan2020pose, burgner2014continuum, franz2014emtracking, proximal_self_sensing}. Marker-based vision instead estimates the end-effector pose from visual markers attached to the instrument~\cite{wu2021closedloop}. These approaches can provide end-effector feedback but require instrument-specific sensing hardware or attached markers.

Markerless vision avoids instrument-mounted sensing and estimates the distal pose from camera images. Wang \textit{et al.} estimated the pose of a continuum surgical instrument by aligning its kinematic model with a random-forest-based end-effector segmentation~\cite{wang2023}. Zhou \textit{et al.} jointly predicted end-effector segmentation and a partial pose, completed the 6D pose using robot-side axial rotation, and further refined the estimate through iterative render-and-compare optimization~\cite{zhou2024markerless}. Their refined estimator achieved a mean position error of 1.24\,mm and an orientation error of $3.20^\circ$, with the iterative refinement stage requiring approximately 800\,ms per frame.

Existing markerless continuum approaches have predominantly relied on single-view endoscopic images and a limited set of visual features~\cite{wang2023, zhou2024markerless}. Stereo fusion of complementary features has also been investigated for rigid laparoscopic instruments~\cite{stereo_rigid_tool_tracking, stereo_unified_feature2025}. MarUco instead integrates multiple complementary geometric features across calibrated stereo views to estimate the full 6D end-effector pose.

\subsection{Synthetic Data Generation for Continuum Manipulators}
\label{synthetic}

Simulation-driven data generation provides scalable training data with automatically generated ground-truth annotations, while domain randomization can facilitate transfer from synthetic to real images~\cite{tobin2017domainrandomizationtransferringdeep, tremblay2018falling}. Graphics processing unit (GPU)-accelerated simulators combine large-scale scene generation, photorealistic rendering, and automated annotation, and have been used for pose estimation and camera-to-robot calibration in general robotic systems~\cite{lee2020cameratorobotposeestimationsingle, lu2023ctrnet}. Frameworks such as ORBIT, Isaac Lab, and ORBIT-Surgical further support scalable simulation for articulated and surgical robots~\cite{Mittal_2023, nvidia2025isaaclabgpuacceleratedsimulation, yu2024orbitsurgicalopensimulationframeworklearning}. Domain-randomized synthetic pipelines have also been developed for pose estimation of rigid surgical instruments~\cite{yoshimura2021mbapose, barragan2024surgical6d}.

Applying such articulated-robot simulation pipelines to continuum manipulators is less direct because their continuously deformable backbones require an explicit approximation before they can be represented using conventional rigid-link robot descriptions. In earlier continuum vision pipelines, Wang \textit{et al.} generated training labels for random-forest end-effector segmentation using a color-based image-processing procedure~\cite{wang2023}. Zhou \textit{et al.} subsequently introduced a custom domain-randomized synthetic generator that requires no manual annotation~\cite{zhou2024markerless}. Their pipeline uses continuum-tool kinematics to verify feasible configurations and renders the tool using randomized textures and illumination before compositing it with real endoscopic background images. Consistent with the domain-randomization paradigm, the pipeline emphasizes broad appearance variation rather than maximizing the realism of individual synthetic images.

MarUco represents the continuum manipulator as an articulated pseudo-rigid-body URDF in Isaac Sim, enabling it to be modeled within a general-purpose robotic simulation framework while preserving kinematic consistency. The pipeline combines ray-traced photorealistic rendering with domain randomization and automatically generates synchronized stereo images together with part-level masks, keypoints with visibility labels, bounding boxes, and per-component 6D pose annotations.

\subsection{Render-and-Compare-Based Pose Estimation}
\label{sec:rw_refine}
Render-and-compare methods exploit the discrepancy between observations and the projection of a known 3D model at an estimated pose. This principle has been used in two main ways. First, differentiable rendering can provide a self-supervised training signal by comparing rendered geometry with image-derived observations, enabling pose-estimation models to adapt using unlabeled real images~\cite{lu2023ctrnet, lu2025ctrnetx}. Second, render-and-compare can refine an initial pose at inference by repeatedly reducing the rendered--observed discrepancy. This has been implemented through gradient-based optimization~\cite{liang2025differentiable_surg_pose, zhou2024markerless, wang2023} or through learned networks that predict relative pose updates from rendered and observed inputs~\cite{li2018deepim, labbe2020cosypose, gcf2020}.

Both optimization-based and learned refinement methods commonly apply repeated rendering-and-update steps to progressively improve the pose estimate~\cite{li2018deepim, zhou2024markerless}. Although effective for pose alignment, this iterative structure introduces additional computation and latency when pose estimates are required continuously for closed-loop feedback.

MarUco uses render-and-compare differently in its offline and online stages. Offline, iterative geometric alignment between predicted and rendered masks and keypoints generates pseudo-ground-truth poses from unlabeled real images for self-supervised adaptation (\Cref{sec:adaptation}). Online, a learned refinement module compares the predicted and rendered geometric features at the initial pose and directly regresses the residual pose correction in a single feedforward pass, avoiding iterative optimization (\Cref{sec:refine_module}).

\section{Hardware and Kinematics}

The experimental platform is a two-segment cable-driven continuum manipulator that adopts the slit-type flexure mechanism reported in~\cite{park2024hysteresis, park2024sam}.

\begin{figure}[t!]
 \centering
 \includegraphics[width=0.72\linewidth]{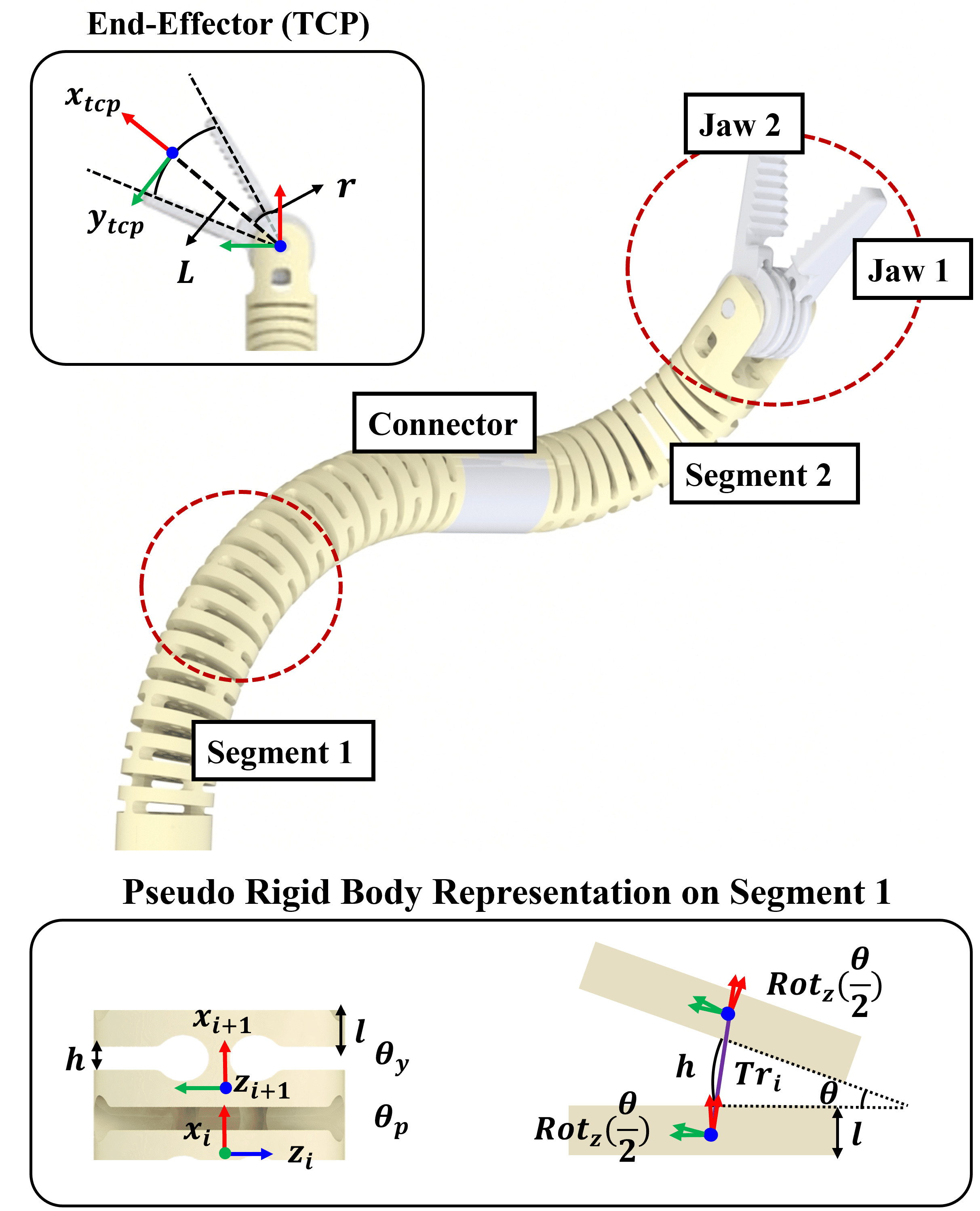}
 \caption{Pseudo-rigid-body (PRB) kinematic model of the two-segment continuum manipulator. (Top) The system includes two flexible segments, a rigid connector, and a dual-jaw gripper. In the PRB model, each flexible segment is discretized into revolute joints and rigid links. The first segment bends in pitch and yaw, the second in pitch only, and the jaw geometry defines the tool center point (TCP) with its equivalent gripper yaw $r$. (Bottom) PRB representation of a slit pair on Segment~1: link length $l$, joint height $h$, and half-angle rotation--translation decomposition of \Cref{eq:t_module}.}
 \label{fig:kinematics_prb}
\end{figure}

The manipulator comprises two flexible segments, a gripper hinge, and a dual-jaw gripper on a central shaft (\Cref{fig:kinematics_prb}), providing seven degrees of freedom (DoF). Ten $2.5$~m-long actuation cables route through an approximately $1.5$~m-long flexible insertion tube to remotely located motors.  The axial translation $q_1$ and global axial rotation $q_2$ move the entire structure. The first flexible segment bends in pitch ($q_3$) and yaw ($q_4$), the second segment bends in pitch only ($q_5$), and the gripper carries two independently actuated yaw joints $q_6$ and $q_7$.
The TCP lies on the bisector axis of the gripper, midway between the two jaw tips along the arc they trace about the hinge (\Cref{sec:gt_annotation}).
The stereo vision system consists of two Basler acA2040-120uc cameras, and all pose-estimation inference was performed on a single NVIDIA RTX A5000 GPU.

\subsection{Forward Kinematics}
\label{sec:fk}
 We model the continuum structure with a pseudo-rigid-body (PRB) representation, discretizing each flexible segment into rigid links connected by revolute joints, with one joint per slit. In the first segment, each slit bends about a single axis, and adjacent slits are rotated $90^\circ$ so that the pitch and yaw slits alternate along the backbone. 
Each consecutive pitch--yaw pair forms  a PRB module that reproduces the  2-DoF bending  of the segment. The homogeneous transformation of a single module is given by

\begin{equation}
\begin{aligned}
\mathbf{T}_{\text{module1}} =
&\;\mathbf{R}_z\!\left(\frac{\theta_p}{2}\right)
\mathbf{T}_x(Tr_i)
\mathbf{R}_z\!\left(\frac{\theta_p}{2}\right)
\mathbf{R}_x\!\left(-\frac{\pi}{2}\right) \\
&\;\mathbf{R}_z\!\left(\frac{\theta_y}{2}\right)
\mathbf{T}_x(Tr_{i+1})
\mathbf{R}_z\!\left(\frac{\theta_y}{2}\right)
\mathbf{R}_x\!\left(\frac{\pi}{2}\right),
\end{aligned}
\label{eq:t_module}
\end{equation}

where $\theta_p$ and $\theta_y$  represent the per-module pitch and yaw angles, respectively, obtained by distributing  segment angles $q_3$ and $q_4$ uniformly over  $n$ modules ($\theta_p = q_3/n$, $\theta_y = q_4/n$).
$Tr_i$  represents the bending-induced translation of the $i$-th PRB element.

\begin{equation}
Tr_i = l \cos\!\left(\frac{\theta_i}{2}\right) + \frac{2h}{\theta_i}\sin\!\left(\frac{\theta_i}{2}\right), \quad i=1,\ldots,2n,
\end{equation}
where, for Segment~1, $i$ indexes the $2n$ slit elements (two per PRB module), $\theta_i \in \{\theta_p, \theta_y\}$ represents the bending angle of the corresponding element, and $l$ and $h$  represent the link length and  joint height of the discretized segment shown in \Cref{fig:kinematics_prb}. In the straight configuration, the expression takes its limit $Tr_i \to l + h$ as $\theta_i \to 0$.
The second segment bends in a single plane; thus, each PRB module  is reduced to one pitch slit, obtained from~\Cref{eq:t_module} by omitting the yaw transformation.

 In this paper, ${}^{A}\mathbf{T}_{B}\in SE(3)$  represents the pose of frame $B$ expressed in frame $A$, mapping coordinates  from frame $B$ to frame $A$.
The base-to-TCP forward kinematics concatenates all module and connector transformations  as

\begin{equation}
\begin{split}
{}^{\text{base}}\mathbf{T}_{\text{TCP}} ={}&
\mathbf{T}_{\text{trans}}\,
\mathbf{T}_{\text{rot}}\,
\mathbf{T}_{\text{seg1}}\,
\mathbf{T}_{\text{connector}} \\
&\cdot\,
\mathbf{T}_{\text{seg2}}\,
\mathbf{T}_{\text{hinge}}\,
\mathbf{T}_{\text{TCP}}(q_6,q_7),
\end{split}
\end{equation}

where $\mathbf{T}_{\text{seg1}}$ and $\mathbf{T}_{\text{seg2}}$ represent the full first- and second-segment transforms, respectively (each formed by the ordered product of the per-module transforms of~\Cref{eq:t_module}; with the yaw term omitted in the second segment); $\mathbf{T}_{\text{trans}}$ and $\mathbf{T}_{\text{rot}}$  represent the base translation $q_1$ and axial rotation $q_2$, respectively; $\mathbf{T}_{\text{connector}}$ and $\mathbf{T}_{\text{hinge}}$  represent the fixed connector and gripper-hinge transforms, respectively; and $\mathbf{T}_{\text{TCP}}(q_6,q_7)$  indicates the TCP based on jaw geometry and the two gripper yaw angles.

\subsection{Jacobian and Inverse Kinematics}
\label{sec:ik}
The geometric Jacobian $\mathbf{J} = [\mathbf{J}_v;\,\mathbf{J}_\omega]$ maps the velocities of  joint variables $\mathbf{q} = [q_1, q_2, q_3, q_4, q_5, r]^\top$ to the TCP spatial velocity, where $r = (q_6 + q_7)/2$  represents the equivalent gripper yaw under the symmetric-jaw TCP definition. During servoing, the jaw opening is  fixed at $40^\circ$, and the two jaw commands are recovered from the reduced coordinates as $q_6 = r - 20^\circ$ and $q_7 = r + 20^\circ$; thus, the loop commands only the mean angle $r$.

The columns accumulate along the PRB chain of \Cref{eq:t_module}. The segment angle is distributed uniformly over the $n$ modules ($\theta_p = q_3/n$), so each half-angle slit rotation turns by $q_3/(2n)$ and contributes $1/(2n)$ of a standard revolute column taken about its own slit axis, evaluated at its own position along the chain. Summing these rotational contributions and the bending-induced changes  in module translations $Tr_i$  provides the segment column. 
The insertion, roll, and gripper-yaw columns follow  standard prismatic and revolute constructions.

Inverse kinematics uses a standard Newton--Raphson update with the Jacobian pseudoinverse, $\mathbf{q}_{k+1}=\mathbf{q}_k+\mathbf{J}^{\dagger}(\mathbf{q}_k)(\mathbf{x}_d-\mathbf{f}(\mathbf{q}_k))$,
where $\mathbf{f}(\mathbf{q})$  and $\mathbf{x}_d$ represent the forward kinematics and  desired TCP pose, respectively. The orientation component of the pose error $\mathbf{x}_d-\mathbf{f}(\mathbf{q}_k)$  represents the axis-angle vector of the relative rotation.

 \begin{figure*}[t!]
 \centering
 \includegraphics[width=0.8\linewidth]{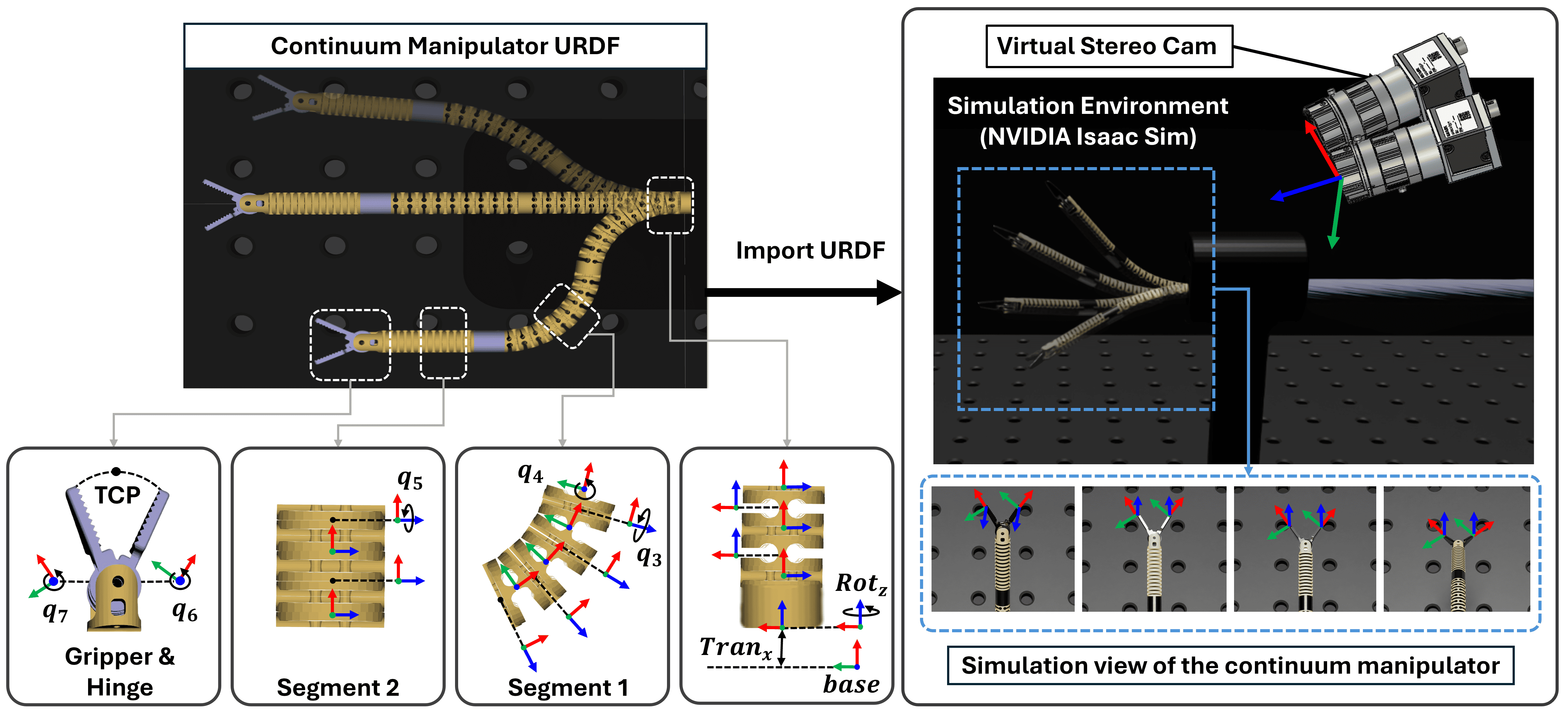}
 \caption{URDF-based modeling and simulation pipeline. (Left) Pseudo-rigid-body URDF discretization of the manipulator, with detailed panels for the base joints ($\mathrm{Tran}_x$ and $\mathrm{Rot}_z$: the axial translation $q_1$ and rotation $q_2$), two segments ($q_3$--$q_5$), and gripper ($q_6$, $q_7$). (Right) Import into Isaac Sim with a calibrated stereo setup for synchronized rendering and automatic pose annotation.}
 \label{fig:urdf}
\end{figure*}

\section{Synthetic Data Generation}
 We constructed a simulation pipeline that renders  photorealistic images in kinematically consistent configurations, automatically generating synchronized stereo pairs with segmentation masks, bounding boxes, keypoints, and 6D pose annotations at scale.

\subsection{URDF Modeling of the Continuum Manipulator}
 We discretize the continuous backbone into a custom URDF of slit-type revolute joints, each following the PRB formulation  in \Cref{sec:fk} (\Cref{fig:urdf}). The slit count  matched the physical flexure design~\cite{park2024sam, park2024hysteresis}. The first segment  included 22 slits alternating between pitch and yaw, grouped into $n=11$ PRB modules of \Cref{eq:t_module}, and the second comprised 11 pitch-only slits ($n=11$).
With the bending limit  in each direction set to $\pm 60^\circ$, the 11 slits  in that direction span  approximately $5.5^\circ$ each, and the segment bending angle is the cumulative rotation $\theta_{\text{segment}}=\sum_{k=1}^{n_{\text{slit}}}\theta_k$ across  the slits. Accordingly, each slit joint in the URDF is assigned a limit of $\pm 0.1$~rad. 
A prismatic joint ($0$--$30~\mathrm{mm}$ stroke)  modeled the axial insertion, and the base rotation and gripper yaw  were independent revolute joints (\Cref{fig:urdf}).
The model  was imported directly into NVIDIA Isaac Sim with kinematically consistent articulation and  photorealistic rendering.
\subsection{Simulation Environment}
 \label{sec:sim_setup}
Synthetic data were generated using NVIDIA Isaac Sim~4.0.0, which combines GPU-accelerated PhysX articulation simulation with ray-traced photorealistic rendering (\Cref{fig:urdf}).  The joint limits and sampling constraints of \Cref{sec:dataset_summary} ensure that every rendered configuration is kinematically valid  with no self-penetration among the densely spaced slits.

Actuation nonlinearities such as hysteresis and friction are not explicitly modeled in the simulator. Because the pose estimator maps camera observations directly to the distal pose rather than actuation commands to robot configuration, the simulator is used primarily to generate kinematically valid configurations together with their corresponding visual observations and pose labels. We therefore sample configurations spanning the operating range described in \Cref{sec:dataset_summary}.

Rendering assigns physically based materials under illumination that combines  spherical area light with a directional source. The jaws received a steel template with tunable metallic and roughness parameters, and the backbone  received domain-randomized diffuse materials (\Cref{sec:domain_rand}).

A virtual stereo RGB pair replicates  real camera parameters (focal length: $12.465~\mathrm{mm}$, resolution: $2048 \times 1536$~px,  and baseline: $63.3~\mathrm{mm}$). Physics and rendering are  synchronously stepped at $60~\mathrm{Hz}$ with gravity disabled. Each step  applied the next sampled configuration and  captured the frame. Each frame records synchronized stereo images, segmentation masks, keypoints, bounding boxes, and the $SE(3)$ poses of Jaw~1, Jaw~2, and the hinge in the left camera frame.

\subsection{Simulation-Driven Ground-Truth Generation}
\label{sec:gt_annotation}
 Supervision was performed within the simulation without manual annotation, producing geometry-consistent keypoints, visibility labels, segmentation masks, bounding boxes, and 6D poses for every frame.

 \begin{figure*}[t!]
 \centering
 \includegraphics[width=0.72\linewidth]{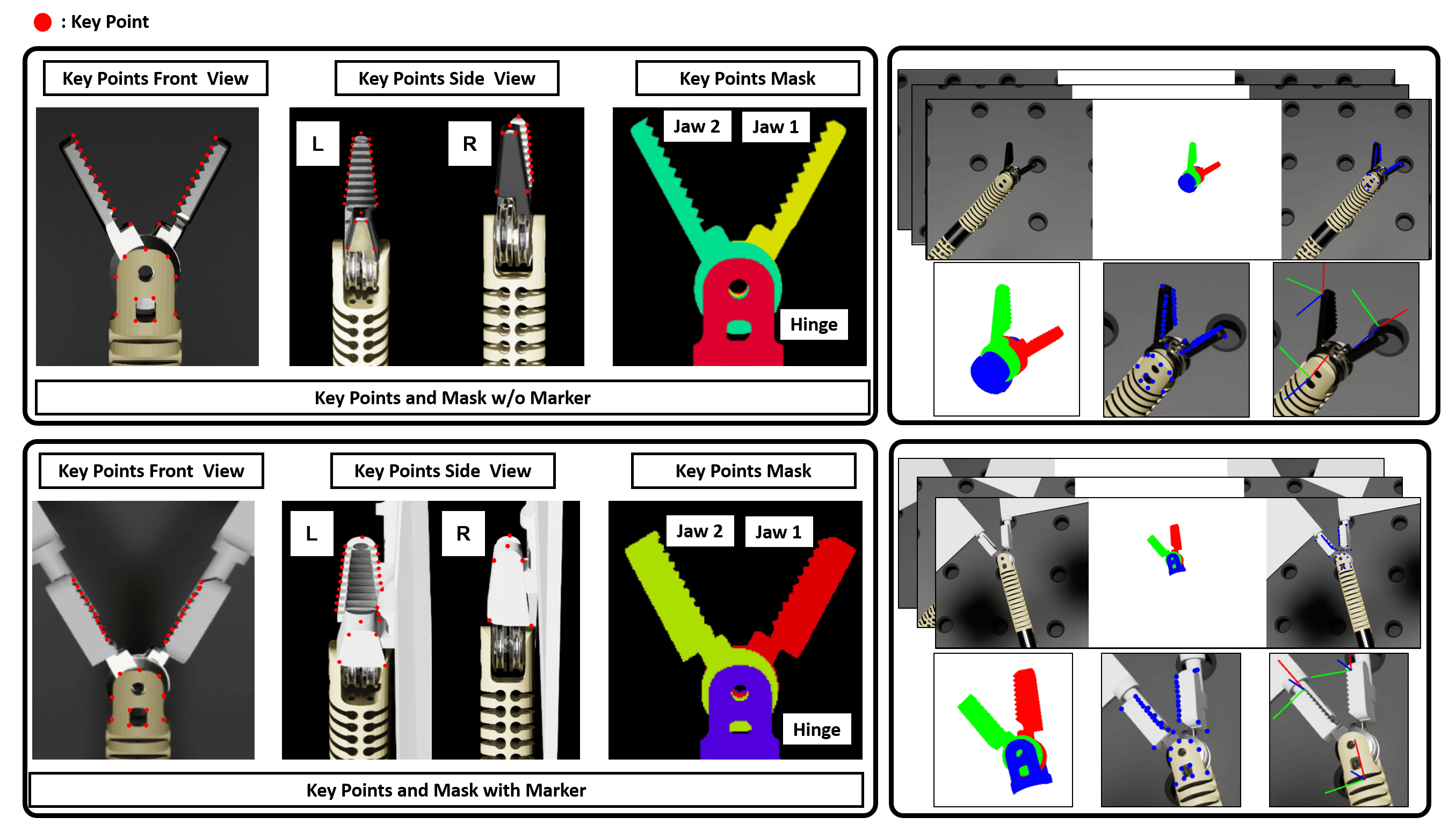}
 \caption{Ground-truth annotation. (Left) Marker-free keypoint and mask definitions (65 keypoints, four-class masks (Jaw~1, Jaw~2, hinge, and background)) and marker-based configuration for real-world ground truth. (Right) Synthetic stereo pairs with automatically generated masks, keypoints, and 6D pose labels. Self-occluded keypoints are excluded by the rendering-based visibility check of \Cref{sec:gt_annotation}.}
 \label{fig:dataset_kpts_mask}
\end{figure*}

\paragraph{Annotation Scope.}
Masks and keypoints cover only the distal components (the two jaws and the hinge) rather than the full backbone because these components directly determine the TCP regulated by the controller. In the experimental setting considered here, the distal assembly remains visible during manipulation, whereas proximal segments may leave the camera view. Prior markerless continuum pose-estimation methods have similarly focused on the distal region~\cite{zhou2024markerless, wang2023}.

\paragraph{Keypoints and Visibility.}
Sparse keypoints complement the dense masks. A mask alone leaves the orientation of each component ambiguous, and keypoints, each fixed to the component body, supply this missing orientation information~\cite{stereo_unified_feature2025}. A total of 65 keypoints captured the distal geometry: 27 along each jaw and 11 on the hinge (\Cref{fig:dataset_kpts_mask}), each defined in the local body frame and projected through the camera model.
As the manipulator articulates, its  geometry can occlude some keypoints from the camera.
Supervising such occluded keypoints introduces label noise; thus, each keypoint carries a visibility flag $\rho_k \in \{0,1\}$.
We compute this flag by rendering the camera-visible surface as a point cloud. A keypoint is labeled as visible ($\rho_k = 1$) when its 3D position lies within $0.1$ mm of the nearest cloud point. The annotation for each keypoint is thus its image coordinates and visibility $\{x_k, y_k, \rho_k\}$, and $\rho_k$ masks the keypoint loss  such that the occluded points are not supervised.

\paragraph{Segmentation Masks and Bounding Boxes.}
Pixel-wise semantic segmentation assigns each pixel to one of  the four classes: background, Jaw~1, Jaw~2, or Hinge (\Cref{fig:dataset_kpts_mask}). Each frame also carries a bounding box that localizes the manipulator and defines the region of interest (ROI) passed to the pose network. The box is  automatically generated by centering the ROI at the image projection of the simulated hinge position.

\paragraph{6D Pose Annotation.}
The simulator provides the 6D poses ${}^{camL}\mathbf{T}_{jaw1}$, ${}^{camL}\mathbf{T}_{jaw2}$, and ${}^{camL}\mathbf{T}_{hinge}\in SE(3)$ in the left camera frame (\Cref{fig:dataset_kpts_mask}). 

The TCP pose for evaluation and control  was analytically reconstructed from the dual-jaw geometry. With each jaw's approach direction $\mathbf{u}_i=\mathbf{R}_{jaw_i}\mathbf{e}_x$ ($\mathbf{e}_x=[1,0,0]^\top$), the opening angle and TCP orientation (the  Jaw~1 frame rotated halfway toward  Jaw~2 about the hinge axis $\mathbf{e}_z$) are

\begin{equation}
\theta = \cos^{-1}(\mathbf{u}_1^\top \mathbf{u}_2),
\qquad
\mathbf{R}_{tcp} = \mathbf{R}_{jaw1}\mathbf{R}_z\!\left(\tfrac{\theta}{2}\right).
\end{equation}

Data generation assigns the jaw indices under a fixed convention, in which $\mathbf{R}_z(\theta/2)$ rotates  Jaw~1 toward  Jaw~2; thus, the unsigned angle determines the TCP orientation without ambiguity.
The TCP position is obtained as

\begin{equation}
\mathbf{t}_{tcp} = 
\frac{\mathbf{t}_{jaw1}+\mathbf{t}_{jaw2}}{2} 
+ 
\mathbf{R}_{tcp}\mathbf{e}_x \, d\!\left(1-\cos\frac{\theta}{2}\right),
\end{equation}

where $d$  represents the known jaw length  and each jaw frame is located at its jaw tip. The first term is the midpoint of the chord connecting the two tips, at a distance $d\cos\frac{\theta}{2}$ from the hinge along the bisector. The correction term $d\left(1-\cos\frac{\theta}{2}\right)$ displaces this midpoint along the bisector onto the arc of radius $d$ traced by the tips, so the TCP lies on this arc midway between the two tips.

\subsection{Domain Randomization}
\label{sec:domain_rand}
 Domain randomization spans three axes.  The backgrounds are either randomly generated solid colors or real endoscopic scenes sampled from a pool of 500 frames~\cite{surgical_open_dataset}. The diffuse color of the backbone  was uniformly randomized in RGB space to prevent overfitting  of the appearance of the prototype. The metallic coefficient of the jaws is sampled uniformly in $[0.4,1.0]$ under physically based shading to vary their specular response.

\subsection{Dataset Composition}
\label{sec:dataset_summary}
 Ablation studies (\Cref{sec:ablation})  used a separate nonrandomized set of 30{,}000 pairs (24{,}000/6{,}000 train/val),  whereas the full deployment pipeline (\Cref{sec:real_validation})  used all 200{,}000 domain-randomized pairs. Each sample  comprised the stereo RGB images, and per view, the segmentation mask, bounding box, $K{=}65$ keypoints with visibility flags, and the $SE(3)$ poses of  Jaw~1,  Jaw~2, and the hinge in the left camera frame. Joint-space sampling spans the operating ranges $q_1 \in [0,10]$~mm, $q_2 \in [-45^\circ,45^\circ]$, and $q_{3\text{--}7} \in [-60^\circ,60^\circ]$, with relative jaw angles constrained to prevent self-intersection.

 \begin{figure*}[t!]
 \centering
 \includegraphics[width=0.8\linewidth]{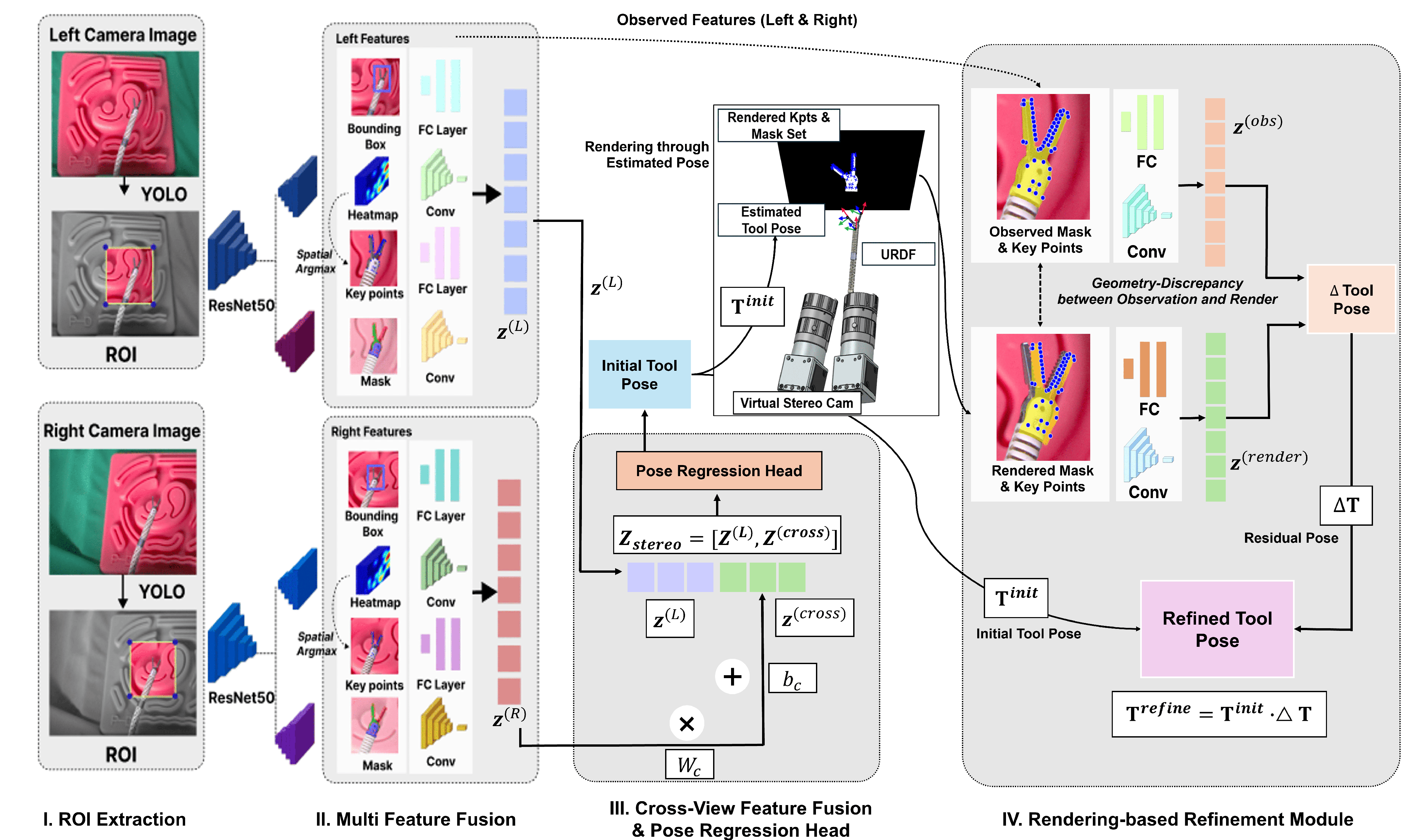}
 \caption{Overview of the proposed pose estimation framework with cross-view feature fusion. (I) ROI extraction: A you only look once (YOLO) detector localizes the manipulator in the left and right stereo images. (II) Multifeature fusion: A shared ResNet-50 encoder extracts keypoints, heatmaps, segmentation masks, and refined bounding boxes, which are embedded and concatenated for pose regression. (III) Cross-view feature fusion \& pose regression head: The right-view embedding is projected by a learned linear map and fused with the left-view embedding, while keeping the output in the left camera frame. (IV) Rendering-based refinement: The initial pose is rendered, and geometric discrepancies between the rendered and observed features are used to predict residual pose corrections in a single feedforward pass.}
 \label{fig:pose_net}
\end{figure*}

\section{Pose Estimation Framework}
 As shown in \Cref{fig:pose_net}, an MFFN jointly predicts keypoints, masks, and heatmaps and regresses an initial 6D pose. A refinement module then renders that pose, compares the rendering with the MFFN observations, and predicts the residual correction in a single pass.

\subsection{Region-of-Interest Extraction}
A YOLO-based detector localizes the manipulator and outputs a bounding box
\(
\mathbf{B}=[x_{\min},y_{\min},x_{\max},y_{\max}]
\)
that crops the input image before feature extraction.
Cropping suppresses background clutter and reduces computation, which  is important in minimally invasive  scenarios where specular highlights and tissue deformation degrade feature learning.

\subsection{Multifeature Fusion Network}
\label{sec:mffn}
 The MFFN jointly estimates dense semantic segmentation, sparse keypoints, heatmaps, and refined spatial extents. These complementary modalities provide global part-level geometry together with fine-grained landmark information for pose regression. 
The network outputs the initial poses of  Jaw~1,  Jaw~2, and the hinge in the left camera frame  with the predicted geometric features $\boldsymbol{\Omega}_{ref}=\{\mathbf{\hat{S}},\mathbf{\hat{K}}\}$, computed for each stereo view ($\boldsymbol{\Omega}^{(L)}_{ref}$  and $\boldsymbol{\Omega}^{(R)}_{ref}$), where $\mathbf{\hat{S}}$  represents segmentation masks and $\mathbf{\hat{K}}$ keypoint coordinates with visibility.

\begin{equation}
\boldsymbol{\Omega}_{ref},\;{}^{camL}\hat{T}_{jaw1,2,hinge}^{(init)}
=
\mathrm{MFFN}(\tilde{I}_L,\tilde{I}_R).
\end{equation}

The monocular baseline used for comparison omits the right view and  considers only $\tilde{I}_L$.

\subsection{Keypoint Detection Module}
 The keypoint branch  predicted $K=65$ geometric landmarks distributed over the distal assembly (\Cref{sec:gt_annotation}).

Given a shared ResNet-50 feature tensor
\(
\mathbf{F}\in\mathbb{R}^{2048\times u_f\times v_f},
\)
an upsampling decoder produces heatmaps
\(
\mathbf{H}\in\mathbb{R}^{K\times u_k\times v_k},
\)
where $u\times v$ denotes the resolution of the cropped ROI input, with $u_f=u/8$, $v_f=v/8$ and $u_k=u/4$, $v_k=v/4$.
Differentiable spatial soft-argmax over the heatmaps $\mathbf{H}_k$ yields continuous keypoint coordinates $(\hat{x}_k,\hat{y}_k)$ as the spatial-softmax expectation~\cite{levine2016end}.
Each keypoint is additionally assigned a predicted visibility confidence
\(
\hat{\rho}_k\in[0,1]
\)
from a sigmoid head, representing reliability under occlusion.

\subsection{Segmentation Module}
 The segmentation branch predicts a four-class semantic mask
\(
\mathbf{\hat{S}}\in\mathbb{R}^{4\times H\times W}
\)
corresponding to  the background, Jaw~1,  Jaw~2, and hinge regions. This part-aware decomposition enables the pose head to disentangle per-component rigid transformations rather than  estimating a single monolithic pose.
A DeepLabv3-style ASPP decoder maps the shared features $\mathbf{F}$ to the low-resolution part probabilities, which are upsampled to full resolution  via bilinear interpolation, $\mathbf{\hat{S}}=\mathrm{Interp}(\mathcal{C}_{seg}(\mathbf{F}))$.

 \begin{figure*}[t!]
 \centering
 \includegraphics[width=0.72\linewidth]{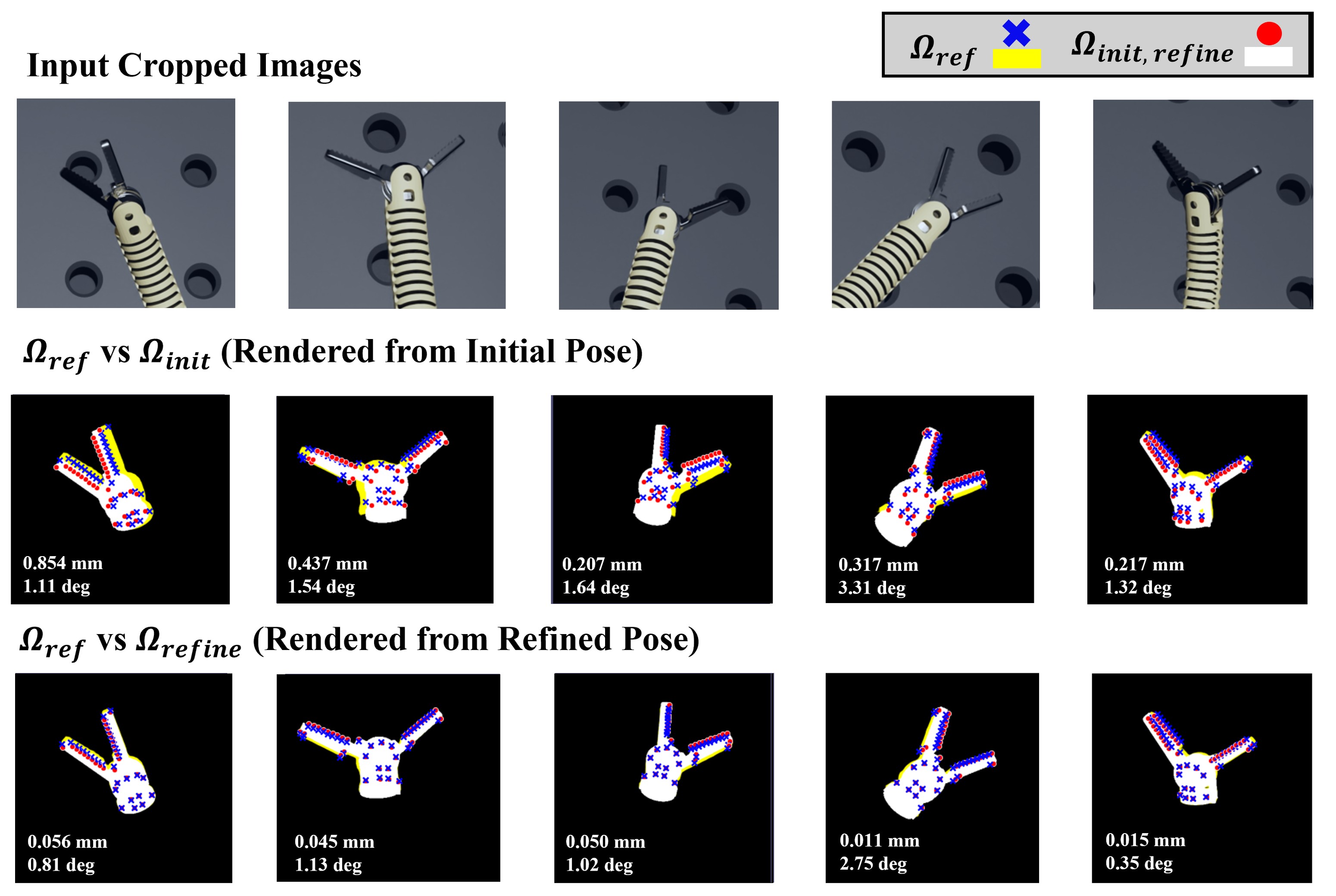}
 \caption{Rendering-based refinement. $\boldsymbol{\Omega}_{ref}$: predicted keypoints (blue) and mask (yellow). $\boldsymbol{\Omega}_{init,\,refine}$: rendered observations $\boldsymbol{\Omega}_{rend}$ at the initial (second row) and refined (third row) pose, shown as rendered keypoints (red) and silhouettes (white). Refinement closes the geometric discrepancy, improving translation (mm) and rotation (deg) accuracy.}
 \label{fig:refinement_module}
\end{figure*}

 \begin{table*}[t]
\caption{TCP pose estimation on the synthetic dataset under monocular and stereo configurations, with and without the refinement module. The overall translation error is the Euclidean distance in millimeters, and the overall rotation error is the axis--angle magnitude in degrees. The axis-wise columns report the per-axis translation errors and angular deviations of predicted TCP axes, expressed in the left camera frame.}

\setlength{\tabcolsep}{3pt}
\renewcommand{\arraystretch}{1.0}
\centering
\footnotesize
\begin{tabular}{l c c c c c c c c c}
\toprule
\textbf{Method} & \textbf{Config}
& \multicolumn{2}{c}{\textbf{Overall}}
& \multicolumn{3}{c}{\textbf{TCP translation error [mm]}}
& \multicolumn{3}{c}{\textbf{TCP axis error [deg]}} \\
\cmidrule(lr){3-4}\cmidrule(lr){5-7}\cmidrule(lr){8-10}
& & Trans [mm] & Rot [deg] & X & Y & Z
 & $\hat{x}_{tcp}$ & $\hat{y}_{tcp}$ & $\hat{z}_{tcp}$ \\
\midrule

MFFN & Mono
& $0.95 \pm 1.70$ & $1.58 \pm 2.19$
& $0.38 \pm 1.2$ & $0.28 \pm 0.6$ & $0.71 \pm 1.2$
& $1.25 \pm 1.9$ & $1.10 \pm 1.5$ & $1.34 \pm 1.9$ \\

MFFN + Refine & Mono
& $0.57 \pm 0.90$ & $1.40 \pm 1.93$
& $0.10 \pm 0.2$ & $0.21 \pm 0.1$ & $0.54 \pm 0.9$
& $1.10 \pm 1.6$ & $1.00 \pm 1.3$ & $1.21 \pm 1.7$ \\

MFFN & Stereo
& $0.64 \pm 0.35$ & $0.95 \pm 0.43$
& $0.29 \pm 0.2$ & $0.21 \pm 0.1$ & $0.43 \pm 0.3$
& $0.80 \pm 0.4$ & $0.70 \pm 0.4$ & $0.74 \pm 0.4$ \\

MFFN + Refine & Stereo
& \textbf{\boldmath $0.14 \pm 0.10$} & \textbf{\boldmath $0.44 \pm 0.24$}
& \textbf{\boldmath $0.05 \pm 0.1$} & \textbf{\boldmath $0.04 \pm 0.1$} & \textbf{\boldmath $0.12 \pm 0.1$}
& \textbf{\boldmath $0.34 \pm 0.2$} & \textbf{\boldmath $0.34 \pm 0.2$} & \textbf{\boldmath $0.36 \pm 0.2$} \\

\bottomrule
\end{tabular}
\label{tab:pose_comparison_sim}
\end{table*}

\subsection{Pose Regression Head}
\label{sec:pose_head}

The pose regression head estimates the initial 6D poses of  Jaw~1,  Jaw~2, and the hinge in the left camera frame by integrating heterogeneous geometric features without explicit 2D--3D correspondences.  Instead of collapsing all modalities into  a single feature-level representation, we encode each feature into a modality-specific latent embedding and fuse them into a unified space.

Foreground segmentation refines the ROI by computing a mask-derived bounding box

\begin{equation}
\mathbf{\hat{B}}_{refined}=
\big[\min_{(\xi,\eta)\in\mathcal{F}}\xi,\;\min_{(\xi,\eta)\in\mathcal{F}}\eta,\;\max_{(\xi,\eta)\in\mathcal{F}}\xi,\;\max_{(\xi,\eta)\in\mathcal{F}}\eta\big],
\end{equation}

where $\mathcal{F}$  represents the foreground pixel set of the predicted segmentation mask $\mathbf{\hat{S}}$ and $(\xi,\eta)$ represents its pixel coordinates ($\mathcal{M}$ is reserved for the link meshes in \Cref{sec:refine_module}).
The refined ROI serves as an additional geometric feature.
 Dense spatial outputs are embedded by strided convolutional encoders and sparse features by fully connected projections :

\begin{equation}
\begin{aligned}
&\mathbf{z}_{heat}=\phi_{heat}(\mathbf{H}),\quad
\mathbf{z}_{seg}=\phi_{seg}(\mathbf{\hat{S}}),\\
&\mathbf{z}_{kpt}=\phi_{kpt}(\mathbf{\hat{K}}),\quad
\mathbf{z}_{box}=\phi_{box}(\mathbf{\hat{B}}_{refined}),
\end{aligned}
\label{eq:encode_kpts}
\end{equation}

with $\phi_{kpt}(\cdot)$ consuming visibility-weighted keypoint coordinates, and the modality embeddings are fused by concatenation : $\mathbf{z}_{fuse}=[\mathbf{z}_{seg},\mathbf{z}_{heat},\mathbf{z}_{kpt},\mathbf{z}_{box}]$.
Regression heads predict the initial poses of  Jaw~1,  Jaw~2, and the hinge  as
\begin{align}
\hat{\mathbf{t}}_i^{(\mathrm{init})} &= g_i^{(t)}(\mathbf{z}_{\mathrm{fuse}}), \nonumber\\
\hat{\mathbf{q}}_i^{(\mathrm{init})} &= g_i^{(r)}(\mathbf{z}_{\mathrm{fuse}}),
\qquad i \in \{\text{jaw1}, \text{jaw2}, \text{hinge}\},
\label{eq:pose_head}
\end{align}

where $g^{(t)}_i(\cdot)$ and $g^{(r)}_i(\cdot)$  represent the component-specific fully connected regression heads for translation and rotation, respectively.
The predicted pose of each component is ${}^{camL}\hat{T}_{i}^{(init)}=[\hat{\mathbf{t}}_i^{(init)},\hat{\mathbf{q}}_i^{(init)}]$, with $\hat{\mathbf{t}}_i^{(init)}\in\mathbb{R}^3$ and unit quaternion $\hat{\mathbf{q}}_i^{(init)}\in\mathbb{S}^3$.

\subsection{Cross-View Feature Fusion}
\label{sec:stereo_aware}
 For stereo, the modality fusion of \Cref{sec:pose_head} is computed independently for each view; thus, the left and right ROIs yield their own fused embeddings $\mathbf{z}^{(L)}$ and $\mathbf{z}^{(R)}$. The two embeddings are combined  using lightweight learned fusion  without explicit triangulation. The right-view embedding is passed through a learned linear projection and concatenated with the left-view embedding.

\begin{equation}
\mathbf{z}^{(cross)} = W_c\,\mathbf{z}^{(R)} + \mathbf{b}_c,
\qquad
\mathbf{z}_{stereo}
=
[\mathbf{z}^{(L)},\,\mathbf{z}^{(cross)}],
\end{equation}

where $W_c$ and $\mathbf{b}_c$ are  jointly trained using the pose heads. The pose heads of \Cref{eq:pose_head}  take $\mathbf{z}_{stereo}$  instead of the single-view $\mathbf{z}_{fuse}$  and learn the cross-view combination directly from the concatenated embedding.
 This asymmetry follows the output convention. All poses are expressed in the left camera frame; therefore, the left embedding is kept as  a reference, and the right view enters through the projection.

\subsection{Single-Pass Rendering-Based Refinement Module}
\label{sec:refine_module}

We address the consistency--latency trade-off discussed in
\Cref{sec:rw_refine} using a learned refinement module that replaces iterative
pose updates with a single feedforward correction. The module renders the CAD
model at the initial pose, compares the rendered geometric features with the
MFFN predictions, and directly regresses the residual pose correction.

The refinement module operates on part-level masks and keypoints rather than
RGB appearance. Consequently, the refinement stage compares rendered and
observed geometry directly without requiring pixel-level matching between
synthetic and real image appearance. Its accuracy nevertheless depends on the
quality of the masks and keypoints predicted by the MFFN.

Given the initial pose estimates
\(
{}^{camL}\hat{T}_i^{(init)},
\; i\in\{\text{jaw1},\text{jaw2},\text{hinge}\},
\)
a differentiable renderer synthesizes geometric observations from the known
CAD meshes:

\begin{equation}
\boldsymbol{\Omega}^{(L,R)}_{rend}
=
\mathrm{Render}\!\left(
{}^{camL}\hat{T}^{(init)}_{jaw1,2,hinge},
\mathcal{M},
{}^{cam(L,R)}T_{base}
\right).
\end{equation}

Here, $\mathcal{M}$ denotes the articulated link meshes. The renderer is
implemented in a base-anchored scene, and the calibrated camera-to-base
extrinsics ${}^{cam(L,R)}T_{base}$ specify the stereo camera poses within this
scene. The predicted component poses are therefore rendered consistently in
both stereo views, with the right-camera observation additionally constrained
by the calibrated stereo geometry.

The rendered features and the MFFN-predicted masks and keypoints
$\boldsymbol{\Omega}^{(L,R)}_{ref}$ are embedded using modality-specific
encoders:

\begin{equation}
\mathbf{z}_{\text{init}}
=
\Phi_{\text{init}}(\boldsymbol{\Omega}^{(L,R)}_{rend}),
\qquad
\mathbf{z}_{\text{obs}}
=
\Phi_{\text{obs}}(\boldsymbol{\Omega}^{(L,R)}_{ref}).
\end{equation}

A lightweight refinement network then estimates the residual correction,

\begin{equation}
\big\{[\Delta\hat{\tilde{\mathbf{t}}}_i,\Delta\hat{\mathbf{q}}_i]\big\}_{i}
=
r_{\phi}\!\left(
\mathbf{z}_{\text{obs}},
\mathbf{z}_{\text{init}}
\right),
\end{equation}

where $r_{\phi}(\cdot)$ denotes the feedforward correction network and
$\Delta\hat{\mathbf{t}}_i = \Delta\hat{\tilde{\mathbf{t}}}_i/s$
is the translation correction applied to the initial pose, with the scaling
factor $s$ defined in \Cref{sec:loss_training}. The refined
pose is obtained by residual composition:

\begin{align}
\hat{\mathbf{t}}_i^{(refine)}
&=
\hat{\mathbf{t}}_i^{(init)}
+
\Delta\hat{\mathbf{t}}_i,\\
\hat{\mathbf{q}}_i^{(refine)}
&=
\hat{\mathbf{q}}_i^{(init)}
\otimes
\Delta\hat{\mathbf{q}}_i,
\end{align}

where $\otimes$ denotes quaternion multiplication. This yields the refined
transformation
${}^{camL}\hat{T}_i^{(refine)}
=
[\hat{\mathbf{t}}_i^{(refine)},\hat{\mathbf{q}}_i^{(refine)}]$.
The training procedure is described in \Cref{sec:loss_training}.

 \begin{figure}[t]
 \centering
 \includegraphics[width=0.72\linewidth]{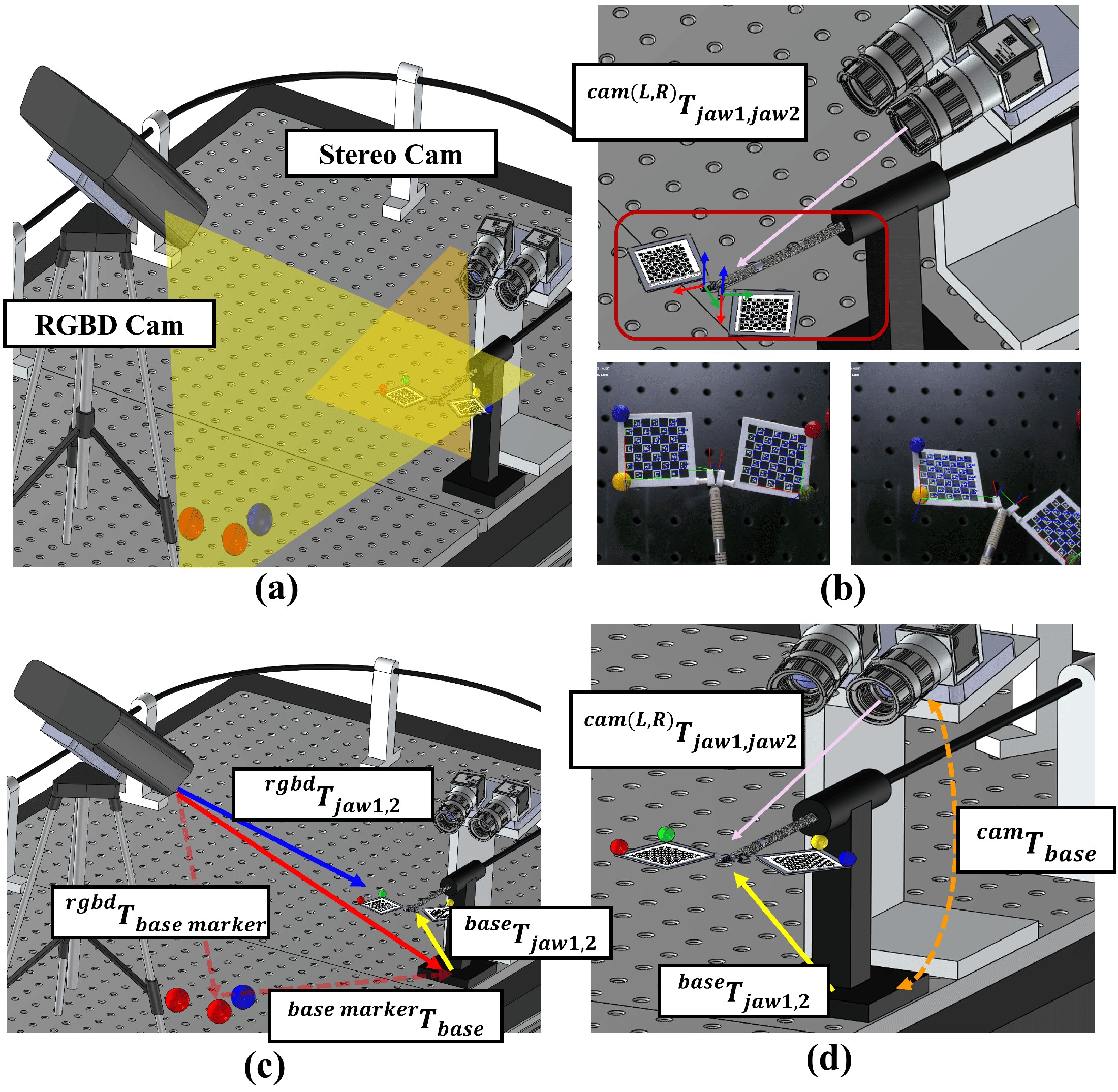}
 \caption{Hand--eye calibration for the camera-to-base extrinsics. (a)~Overall setup. (b)~Stereo estimation of the camera-to-jaw poses from ChArUco markers. (c)~RGB-D localization of the jaw fiducials by RANSAC sphere fitting; combined with the RGB-D-to-base transformation pre-calibrated from a base-mounted marker, this yields the base-to-jaw poses without forward kinematics. (d)~Chaining (b) and (c) gives per-configuration extrinsic candidates, which are jointly optimized into final extrinsics.}
 \label{fig:hand_eye_cal}
\end{figure}

\subsection{Loss Formulation and Multistage Training}
\label{sec:loss_training}

The framework is optimized with stage-specific objectives aligned  with each module,  stabilizing convergence and mitigating gradient interference between perception and 6D pose estimation.

\subsubsection*{Stage 1: Visual Perception Learning}
 The YOLO detector is  independently trained using standard detection losses. The segmentation, keypoint, and visibility branches are jointly optimized  using the composite objective

\begin{equation}
\mathcal{L}_{\mathrm{perc}}
=
\lambda_{\mathrm{seg}}\mathcal{L}_{\mathrm{seg}}
+
\lambda_{\mathrm{kpt}}\mathcal{L}_{\mathrm{kpt}}
+
\lambda_{\mathrm{vis}}\mathcal{L}_{\mathrm{vis}},
\end{equation}

where $\lambda_{\mathrm{seg}} = \lambda_{\mathrm{vis}} = 1$ and $\lambda_{\mathrm{kpt}} = 10^{-2}$; the small keypoint weight compensates for the pixel-scale magnitude of the  smooth $\ell_1$ term.
Segmentation loss combines pixelwise classification with regional overlap.

 \begin{equation}
\mathcal{L}_{\mathrm{seg}}
=
\mathrm{CE}_{w}(\hat{\mathbf{S}},\mathbf{S})
+
\Bigg(
1-
\frac{2\langle \hat{\mathbf{S}},\mathbf{S}\rangle}{\|\hat{\mathbf{S}}\|_1+\|\mathbf{S}\|_1}
\Bigg),
\end{equation}
where $\mathrm{CE}_{w}$  represents the cross-entropy over the four semantic classes (background,  Jaw~1,  Jaw~2, and hinge) with class weights $[0.08, 1, 1, 1]$ that downweight the dominant background class. The second term is the multiclass soft Dice loss.
 Keypoint localization is supervised using a visibility-weighted  smooth $\ell_1$ loss  function.

\begin{equation}
\mathcal{L}_{\mathrm{kpt}}
=
\frac{1}{\sum_{j}\rho_j^{gt}}
\sum_{j=1}^{K}
\rho_j^{gt}\,
\mathrm{SmoothL1}\!\left(
\hat{\mathbf{k}}_j-\mathbf{k}_j
\right),
\end{equation}

where $\rho_j^{gt}\in\{0,1\}$  represents the ground-truth visibility of the $j$-th keypoint, so that occluded landmarks contribute no gradient and the loss averages only over visible landmarks. The predicted visibility serves as  inference time confidence and enters the pseudo-ground-truth objective of \Cref{sec:adaptation}, where no ground truth is available.
Visibility is optimized with a squared error $\mathcal{L}_{\mathrm{vis}}=\|\hat{\boldsymbol{\rho}}-\boldsymbol{\rho}^{gt}\|_2^2$. The keypoint coordinates in $\mathcal{L}_{\mathrm{kpt}}$ are expressed in cropped-ROI pixels.

 \begin{figure}[t]
 \centering
 \includegraphics[width=0.7\linewidth]{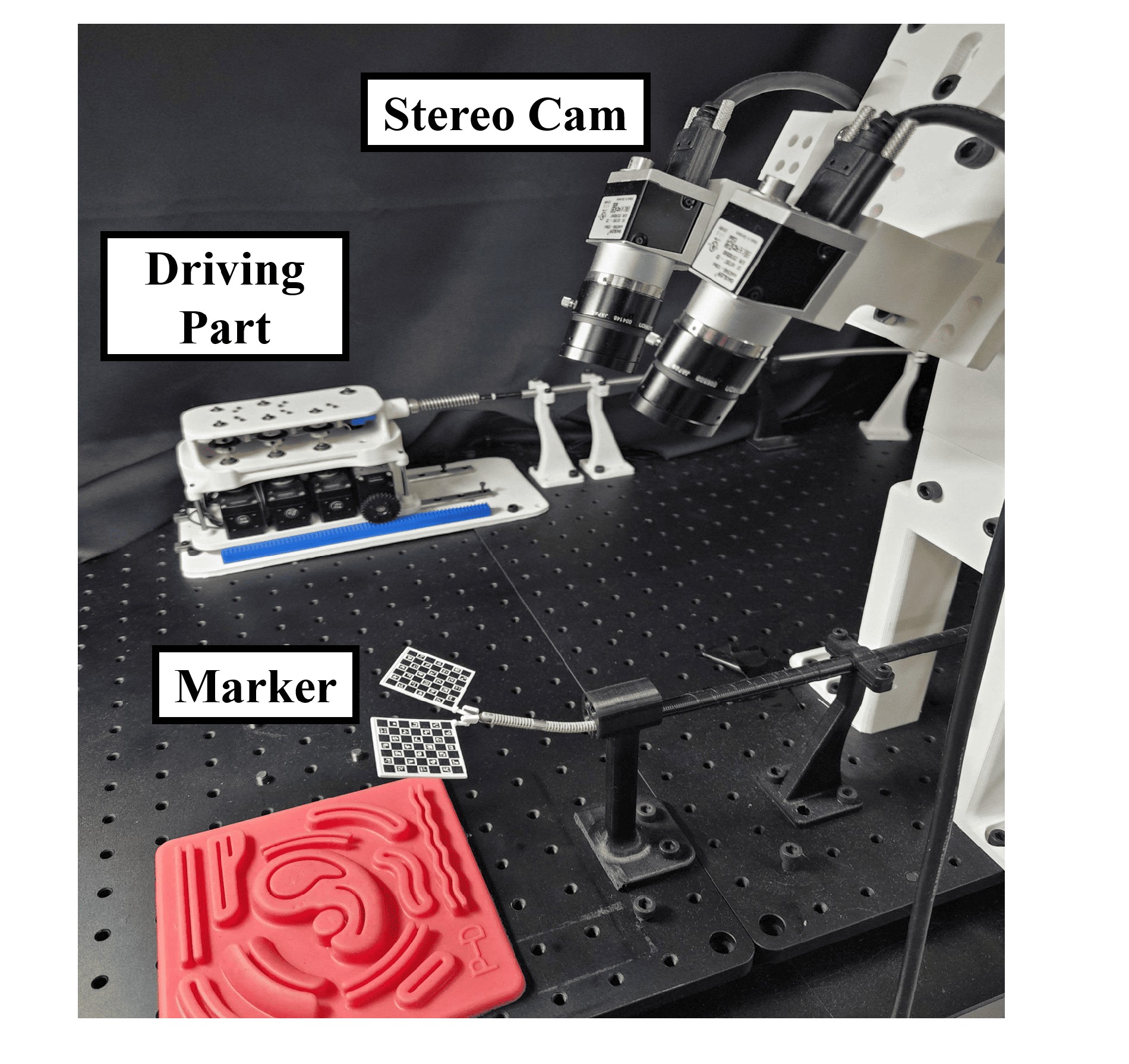}
 \caption{Real-world experimental setup: Actuation unit, two-segment continuum manipulator, stereo camera pair, and jaw-mounted ChArUco cases for ground-truth camera-to-TCP pose. The manipulator is driven by tendons of approximately 2.5\,m routed through a flexible insertion tube of approximately 1.5\,m.}
 \label{fig:real_setup}
\end{figure}

\subsubsection*{Stage 2: Pose Regression Training}
 Stage 2 freezes all perception branches and trains the pose regression head  using a weighted combination of translation and orientation errors.

\begin{equation}
\mathcal{L}_{\mathrm{pose}}
=
\sum_{i}
\lambda_t\, \ell_{\delta}\!\big(\hat{\mathbf{t}}_i^{(init)}-\mathbf{t}_i\big)
+
\lambda_r \big(1-\langle \hat{\mathbf{q}}_i^{(init)},\mathbf{q}_i\rangle^2\big),
\end{equation}

where $\hat{\mathbf{t}}_i^{(init)}\in\mathbb{R}^3$ and $\hat{\mathbf{q}}_i^{(init)}\in\mathbb{S}^3$  represent the translation and unit quaternion predicted by the pose head of \Cref{sec:pose_head} for component $i \in \{\mathrm{jaw1},\mathrm{jaw2},\mathrm{hinge}\}$, and $(\mathbf{t}_i,\mathbf{q}_i)$ represents the corresponding ground-truth pose. The superscript marks this prediction as the initial pose  prior to refinement (\Cref{sec:refine_module}).

The term $\ell_{\delta}$ represents the smooth $\ell_1$ (Huber) loss with a threshold $\delta=10^{-2}$ applied elementwise to the translation error. Translations are represented in millimeters after subtracting a fixed workspace offset for numerical centering. The weights $\lambda_t=0.05$ and $\lambda_r=30$ are shared across the three components. The squared cosine similarity resolves the antipodal ambiguity of the quaternion double cover.

\subsubsection*{Stage 3: Rendering-Based Refinement}

The refinement module is trained with all preceding components frozen, learning the residual that maps the initial prediction to the ground truth rather than the refined pose itself. The residual supervision is constructed as

\begin{align}
\Delta \mathbf{t}_i^{\ast} &= \mathbf{t}_i - \hat{\mathbf{t}}_i^{(\mathrm{init})},
\qquad
\Delta \tilde{\mathbf{t}}_i^{\ast} = s\,\Delta \mathbf{t}_i^{\ast}, \quad s=100, \\\nonumber
\Delta \mathbf{q}_i^{\ast} &= \big(\hat{\mathbf{q}}_i^{(\mathrm{init})}\big)^{-1} \otimes \mathbf{q}_i,
\end{align}

where $(\cdot)^{-1}$ represents quaternion conjugation and $\Delta \tilde{\mathbf{t}}_i^{\ast}$ is the scaled translation target.
The translation head outputs the prediction $\Delta \hat{\tilde{\mathbf{t}}}_i$ in this scaled representation, and the corrections are supervised by the residual loss

\begin{equation}
\mathcal{L}_{\mathrm{ref}}
=
\sum_{i}
\lambda_t^{(\mathrm{ref})}\,
\ell_{\delta}\!\big(\Delta \hat{\tilde{\mathbf{t}}}_i - \Delta \tilde{\mathbf{t}}_i^{\ast}\big)
+
\lambda_r^{(\mathrm{ref})}
\big(1 - \langle \Delta \hat{\mathbf{q}}_i, \Delta \mathbf{q}_i^{\ast} \rangle^2\big).
\end{equation}

At inference, the applied translation correction is recovered as
$\Delta \hat{\mathbf{t}}_i = \Delta \hat{\tilde{\mathbf{t}}}_i / s$ before the residual composition of \Cref{sec:refine_module}.

The loss form and weights are shared with Stage~2 ($\ell_{\delta}$ with $\delta=10^{-2}$, $\lambda_t^{(\mathrm{ref})}=0.05$, and $\lambda_r^{(\mathrm{ref})}=30$). Translation residuals are expressed in millimeters before the scaling by $s$.
Stages~2 and~3 are trained  using the Adam optimizer (learning rates $10^{-3}$ and $10^{-4}$, respectively)  with a batch size of 8, plateau-based learning-rate decay, and early stopping  based on validation loss.
  \begin{algorithm}[t!]
\caption{ Self-supervised real adaptation with  pseudo ground truth refinement}
\label{alg:tool_estimation}
\begin{algorithmic}[1]
 \Require Stereo images $I_L, I_R$, link meshes $\mathcal{M}$, extrinsics
${}^{cam}T_{base}^{L}, {}^{cam}T_{base}^{R}$, multifeature fusion network $f_{\theta}$,
refinement module $r_{\phi}$
\Ensure Adapted multifeature fusion network $f_{\theta^*}$ and refinement module $r_{\phi^*}$
 \State Initialize pseudo-GT buffer $\mathcal{D}_{pseudo} \leftarrow \emptyset$
 \State // Forward prediction from stereo observation
\State $(\Omega^{L,R}_{ref}, \hat{T}) \leftarrow f_{\theta}(I_L, I_R)$
\State $\hat{T} \equiv {}^{camL}T_{jaw1,2,hinge}$, parameterized per component as translation $\mathbf{t}_c$ and quaternion $\mathbf{q}_c$
 \State // Pseudo GT pose refinement with differentiable rendering
\For{$i = 1$ to $N_{opt}$}
 \State $\Omega^{L,R}_i \leftarrow \mathrm{Render}(\hat{T}, \mathcal{M}, {}^{cam(L,R)}T_{base})$
 \State $L_i \leftarrow \mathcal{L}_{align}(\Omega^{L,R}_i, \Omega^{L,R}_{ref})$
 \State $(\mathbf{t}_c, \mathbf{q}_c) \leftarrow (\mathbf{t}_c, \mathbf{q}_c) - \eta\, \nabla_{(\mathbf{t}_c, \mathbf{q}_c)} L_i$ \Comment{$c \in \{\text{jaw1}, \text{jaw2}, \text{hinge}\}$}
 \State $\mathbf{q}_c \leftarrow \mathbf{q}_c / \lVert\mathbf{q}_c\rVert$,\; $\hat{T} \leftarrow \big(\mathbf{t}_c, \mathbf{R}(\mathbf{q}_c)\big)$
\EndFor
 \State // Store refined pseudo-GT sample
\State $\mathcal{D}_{pseudo} \leftarrow \mathcal{D}_{pseudo} \cup \{(I_L, I_R, \hat{T})\}$
 \State Freeze all visual perception branches of $f_{\theta}$
\State // Few-shot real adaptation (pose regression only)
\If{$|\mathcal{D}_{pseudo}| \geq N_{pseudo}$}
 \For{$e = 1$ to $E_{fs}$}
 \State Update pose regression head of $f_{\theta}$ and refinement module $r_{\phi}$ using $\mathcal{D}_{pseudo}$
 \EndFor
\EndIf
 \State Set $f_{\theta^*} \leftarrow f_{\theta}$, $r_{\phi^*} \leftarrow r_{\phi}$
\State \Return $f_{\theta^*}, r_{\phi^*}$
 \end{algorithmic}
\end{algorithm}

\section{Model Validation on Synthetic Data}
\label{sec:ablation}
 The contributions of the refinement module and  cross-view feature fusion  are evaluated on the nonrandomized synthetic subset of \Cref{sec:dataset_summary} (24{,}000 training  and 6{,}000 validation pairs), with all models evaluated in native PyTorch. The analyses on the real dataset are reported in \Cref{sec:real_validation}.

\paragraph{Overall and  axis-wise accuracy.}
\Cref{tab:pose_comparison_sim} summarizes TCP accuracy (translation: Euclidean distance to ground truth; rotation: axis--angle magnitude of $q_{rel} = q_{gt}^{-1} \otimes q_{pred}$; error pairs are written as translation\,/\,rotation throughout). The  full-stereo MFFN + Refine attains $0.14$~mm\,/\,$0.44^\circ$, representing an $85\%$\,/\,$72\%$ reduction relative to the monocular MFFN. The axis-wise columns show that $z$-axis translation errors are consistently larger than $x$ and $y$, indicating greater depth difficulty,  whereas orientation errors are similar across  all three TCP axes.

\paragraph{Effect of the  refinement module.}
\Cref{fig:refinement_module} compares the rendered observations at the initial and refined poses.  Renderings at the initial pose  deviated visibly from the predicted masks and keypoints,  whereas the single-pass correction reduces this discrepancy. Quantitatively, refinement reduces  translation and rotation errors by $40\%$ and $11\%$ in the monocular configuration, and by $78\%$ and $54\%$ in the stereo configuration, acting across all translational axes  (\Cref{tab:pose_comparison_sim}).

\paragraph{Effect of the  stereo configuration.}
 The stereo input yielded complementary improvements of $33\%$ in translation and $40\%$ in rotation for  MFFN (\Cref{tab:pose_comparison_sim}).  Gains are concentrated along the camera depth direction, where the full framework reduces the $z$-axis error from $0.54$ to $0.12$  mm and the $\hat{z}_{tcp}$ orientation error from $1.21^\circ$ to $0.36^\circ$. A single view  infers depth only indirectly, from monocular cues such as mask scale. A stereo pair encodes depth geometrically in the cross-view disparity, and the  depth-axis gains suggest that the fused embedding learns to use this cue.

 \begin{table}[t!]
\caption{Effect of jaw case attachment on TCP pose estimation accuracy in the synthetic dataset. Results compare models trained with and without the jaw case geometry.}
\centering
\setlength{\tabcolsep}{4pt}
\renewcommand{\arraystretch}{1.0}
\begin{tabular}{l c c c}
\hline
\textbf{Method} & \textbf{Jaw Case} 
& \begin{tabular}[c]{@{}c@{}}\textbf{TCP Trans}\\ \textbf{Err [mm]}\end{tabular}
& \begin{tabular}[c]{@{}c@{}}\textbf{TCP Rot}\\ \textbf{Err [deg]}\end{tabular} \\ \hline
MFFN + Refine (Stereo) & Off
& 0.14 $\pm$ 0.10 & 0.44 $\pm$ 0.24 \\
MFFN + Refine (Stereo) & On
& 0.17 $\pm$ 0.12 & 0.53 $\pm$ 0.30 \\ \hline
\end{tabular}
\label{tab:jaw_case_attachment}
\end{table}

 \begin{table}[t!]
\centering
\caption{Processing time and model size of each pipeline component measured over 1{,}000 real stereo pairs.}
\label{tab:time_param}
\setlength{\tabcolsep}{4pt}
\begin{tabular}{lccc}
\toprule
Module & \# Params & PyTorch [ms] & TensorRT [ms]\\
\midrule
YOLO & 56.9M & $33.9 \pm 1.3$ & $6.8 \pm 0.4$ \\
MFFN & 84.7M & $131.5 \pm 17.9$ & $31.4 \pm 3.9$ \\
Rendering (L+R) & -- & $13.5 \pm 2.0$ & $13.4 \pm 0.7$ \\
Refinement Module & 0.84M & $8.2 \pm 2.6$ & $0.8 \pm 0.1$ \\
\midrule
Total & 142.4M & $187.0 \pm 24.4$ & $52.4 \pm 6.0$ \\
\bottomrule
\end{tabular}
\end{table}

\section{Sim-to-Real Transfer and Deployment}
\label{sec:sim2real}

This section details the transfer of the framework to a physical system,
starting with the models trained on $200{,}000$ domain-randomized synthetic
pairs (\Cref{sec:dataset_summary}). First, we calibrate the real stereo cameras
without relying on robot kinematics (\Cref{sec:calibration}) and mitigate
residual systematic pose bias through self-supervised adaptation on unlabeled
real stereo pairs (\Cref{sec:adaptation}). Ground-truth acquisition and
real-world evaluation require jaw-mounted markers, so we assess how the
attached marker cases affect pose estimation (\Cref{sec:jaw_case}). Finally,
the pipeline is compiled using TensorRT for low-latency online deployment
(\Cref{sec:time_efficiency}).

\begin{figure*}[t!]
 \centering
 \includegraphics[width=0.72\linewidth]{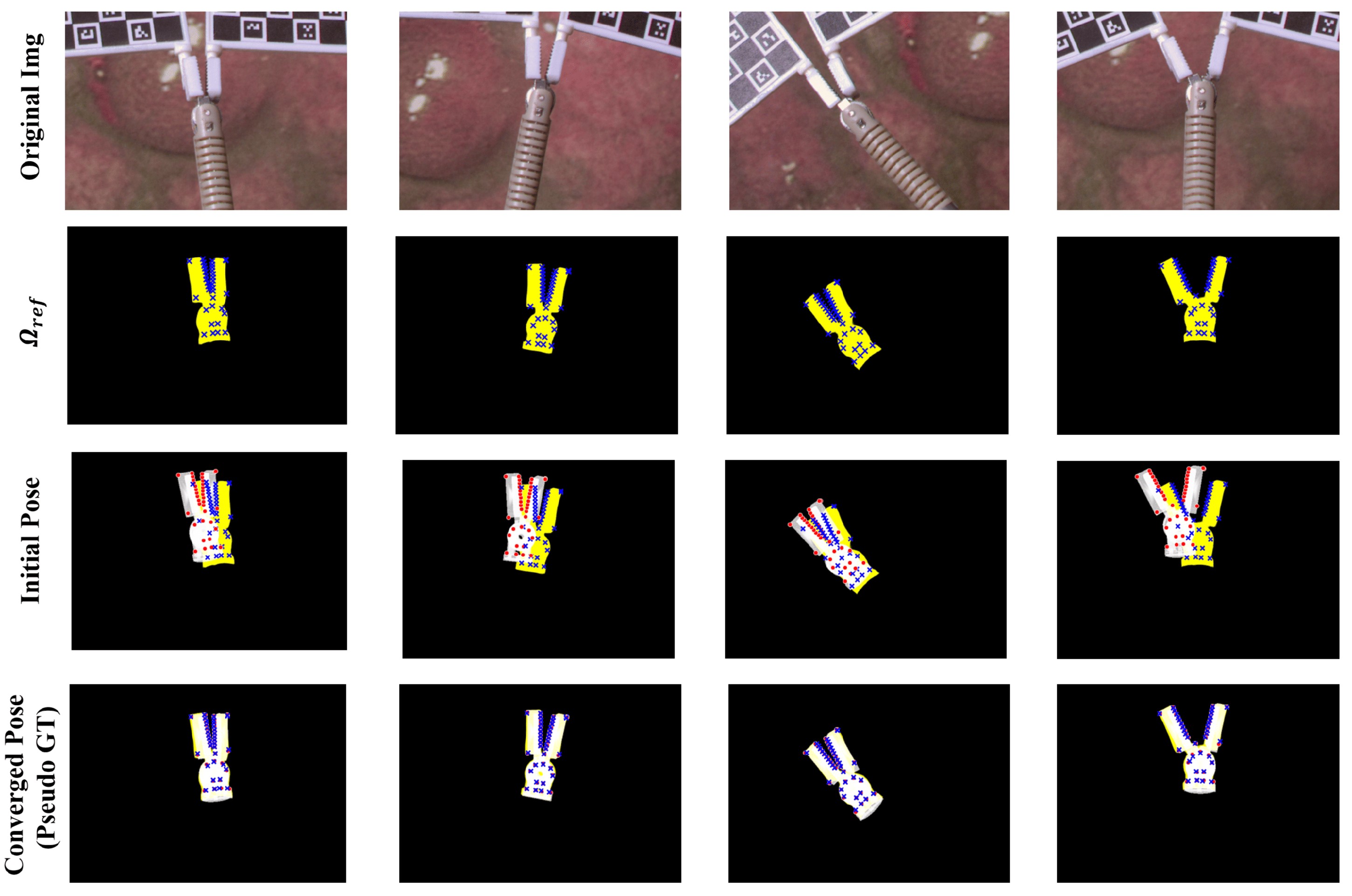}
\caption{Rendering-based pseudo ground-truth generation. From a real image (top), the network predicts keypoints and masks ($\boldsymbol{\Omega}_{ref}$; yellow mask, blue crosses). The initial pose is rendered (third row; gray mesh, red keypoints) and the geometric discrepancy is iteratively minimized; the converged pose (bottom) serves as pseudo ground truth.}
 \label{fig:pseudo_gt}
\end{figure*}

\subsection{Real Camera Intrinsic and Extrinsic Calibration}
\label{sec:calibration}

Camera intrinsics are obtained using standard per-camera checkerboard
calibration. Extrinsics are calibrated using the setup in
\Cref{fig:real_setup}, following \Cref{fig:hand_eye_cal}. Classical hand--eye
formulations typically rely on end-effector poses obtained from accurate
forward kinematics. For the tendon-driven continuum manipulator considered
here, hysteresis makes these kinematic poses unreliable for calibration.
We therefore estimate the extrinsics without requiring kinematically known
end-effector poses.

We employ two independent sensing channels: an RGB-D camera pre-calibrated to
the robot base and the target stereo camera pair. Spherical fiducials in the
jaw case are localized by the RGB-D camera using hue, saturation, and value
(HSV) segmentation and RANSAC point-cloud fitting~\cite{ransac}, while the
stereo system estimates the jaw pose from ChArUco markers~\cite{aruco2014} in
its own frame. For clarity, the jaw index is omitted below. The same procedure
is performed independently for each jaw.

The RGB-D camera is fixed in the workspace, and its RGB-D-to-base
transformation ${}^{rgbd}\mathbf{T}_{base}$ is calibrated once from
base-mounted fiducials and stored. For each configuration $i$, the RGB-D
camera localizes the jaw fiducials to yield
${}^{rgbd}\mathbf{T}^{(i)}_{jaw}$. The stored transformation then provides the
base-to-jaw pose without relying on continuum-manipulator forward kinematics.

\begin{equation}
{}^{base}\mathbf{T}^{(i)}_{jaw}
=
\left({}^{rgbd}\mathbf{T}_{base}\right)^{-1}
{}^{rgbd}\mathbf{T}^{(i)}_{jaw}.
\end{equation}

The stereo system measures the same jaw pose in its own frame as
${}^{cam}\mathbf{T}^{(i)}_{jaw}$. Chaining the two measurements provides one
camera-to-base extrinsic candidate per configuration.

\begin{equation}
{}^{cam}\mathbf{T}^{(i)}_{base}
=
{}^{cam}\mathbf{T}^{(i)}_{jaw}
\left({}^{base}\mathbf{T}^{(i)}_{jaw}\right)^{-1}.
\end{equation}

This procedure yields a set of extrinsic estimates
$\{{}^{cam}\mathbf{T}^{(i)}_{base}\}_{i=1}^{N}$, where $N=400$.
The RGB-D camera is used only during calibration, whereas the deployed system
uses only the stereo cameras. Because the base-to-jaw and camera-to-jaw poses
are obtained from independent sensing channels, the construction does not rely
on the continuum-manipulator forward kinematics. The candidate estimates
exhibit standard deviations of approximately 7--8\,mm and
$1.4$--$1.5^\circ$, reflecting combined sensing and marker-localization
variability, including uncertainty in the manually measured offset between
the base fiducial and robot base. An initial estimate is obtained by averaging
translations in Euclidean space and rotations through quaternion averaging.

We further refine the extrinsics using robust optimization,

\begin{equation}
\min_{\mathbf{X} \in SE(3)}
\sum_i \sum_{k=1}^{6} \varrho \left(
\left[
\boldsymbol{\Lambda} \operatorname{Log} \left(
\mathbf{X}^{-1} {}^{cam}\mathbf{T}^{(i)}_{base}
\right)
\right]_k
\right),
\label{eq:transform_optimize}
\end{equation}

where $\operatorname{Log}\colon SE(3) \to \mathbb{R}^6$ maps the residual
transformation between the current estimate $\mathbf{X}$ and the $i$-th
candidate to its translational and rotational components. The diagonal matrix
$\boldsymbol{\Lambda}$ scales these components to comparable magnitudes, and
the Huber loss $\varrho(\cdot)$ reduces the influence of outlier candidates.
Residual systematic pose bias remaining after calibration is further mitigated
through the self-supervised adaptation described in \Cref{sec:adaptation}.

\subsection{Self-Supervised Real Adaptation with Pseudo Ground Truth}
\label{sec:adaptation}

Residual uncertainty in the estimated camera extrinsics can introduce a
systematic offset between the synthetic training geometry and the physical
stereo system. Together with other sim-to-real discrepancies, this can produce
systematic pose errors when the synthetically trained model is applied to real
images. To adapt the pose estimator without manual pose annotation, we
construct pseudo ground-truth (GT) pose supervision directly from real
observations through differentiable rendering
(\Cref{alg:tool_estimation}).

For each real stereo pair, the MFFN $f_{\theta}$ predicts geometric features
$\boldsymbol{\Omega}^{(L,R)}_{ref}
=\{\hat{\mathbf{S}}^{(L,R)},\hat{\mathbf{K}}^{(L,R)}\}$
and an initial pose ${}^{camL}\hat{T}^{(init)}$. Starting from
$\hat{T}={}^{camL}\hat{T}^{(init)}$, the link meshes $\mathcal{M}$ are
rendered in both views as

\begin{equation}
\boldsymbol{\Omega}^{(L,R)}_{rend}
=
\mathrm{Render}(\hat{T}, \mathcal{M}, {}^{cam(L,R)}T_{base}),
\end{equation}

where
$\boldsymbol{\Omega}^{(L,R)}_{rend}
=\{\mathbf{S}_{rend}^{(L,R)},\mathbf{K}_{rend}^{(L,R)}\}$
contains the rendered silhouettes and keypoints.

\paragraph{Rendering-based alignment loss}
We refine the pose used for pseudo-GT generation by minimizing the
rendering-based alignment loss between the rendered and predicted geometric
features~\cite{liang2025differentiable_surg_pose}:

\begin{equation}
\mathcal{L}_{align}
=
\lambda_1 \mathcal{L}_{MSE}
+ \lambda_2 \mathcal{L}_{dist}
+ \lambda_3 \mathcal{L}_{scale}
+ \lambda_4 \mathcal{L}_{kpt}^{align}.
\end{equation}

The silhouette consistency term is

\begin{equation}
\mathcal{L}_{MSE}
=
\frac{1}{HW}
\sum_{i=0}^{H-1}\sum_{j=0}^{W-1}
\left(
\mathbf{S}_{rend}(i,j)-\hat{\mathbf{S}}_{ref}(i,j)
\right)^2.
\end{equation}

The MSE provides weak gradients when the rendered and predicted silhouettes
do not overlap, so a distance-field term is additionally used:

\begin{equation}
\mathbf{D}_{ref}(i,j)=
\begin{cases}
0, & \hat{\mathbf{S}}_{ref}(i,j)=1,\\
\dfrac{\mathrm{dist}(i,j)}{\gamma}, & \hat{\mathbf{S}}_{ref}(i,j)=0,
\end{cases}
\end{equation}

where $\mathrm{dist}(i,j)$ and $\gamma$ denote the Euclidean distance to the
nearest foreground pixel and the decay factor, respectively. The corresponding
loss is

\begin{equation}
\mathcal{L}_{dist}
=
\sum_{i=0}^{H-1}\sum_{j=0}^{W-1}
\mathbf{S}_{rend}(i,j)\, \mathbf{D}_{ref}(i,j).
\end{equation}

Scale consistency is defined by comparing the normalized foreground areas:

\begin{equation}
\mathcal{L}_{scale}
=
\left|
 \frac{\sum_{i,j} \mathbf{S}_{rend}(i,j)}{\sum_{i,j} 1}
-
 \frac{\sum_{i,j} \hat{\mathbf{S}}_{ref}(i,j)}{\sum_{i,j} 1}
\right|.
\end{equation}

Keypoint alignment uses a visibility-weighted smooth $\ell_1$ loss,

\begin{equation}
\mathcal{L}_{kpt}^{align}
=
\frac{1}{\sum_{j}\hat{\rho}_j}
\sum_{j=1}^{K}
\hat{\rho}_j\,
\mathrm{SmoothL1}
\!\left(
\mathbf{K}_{rend}^{j}-\hat{\mathbf{K}}_{ref}^{j}
\right),
\end{equation}

where $\hat{\rho}_j\in[0,1]$ denotes the predicted visibility confidence of the
$j$-th keypoint, weighting the contribution of each landmark to the alignment
objective.

 \begin{figure*}[t!]
 \centering
 \includegraphics[width=0.9\linewidth]{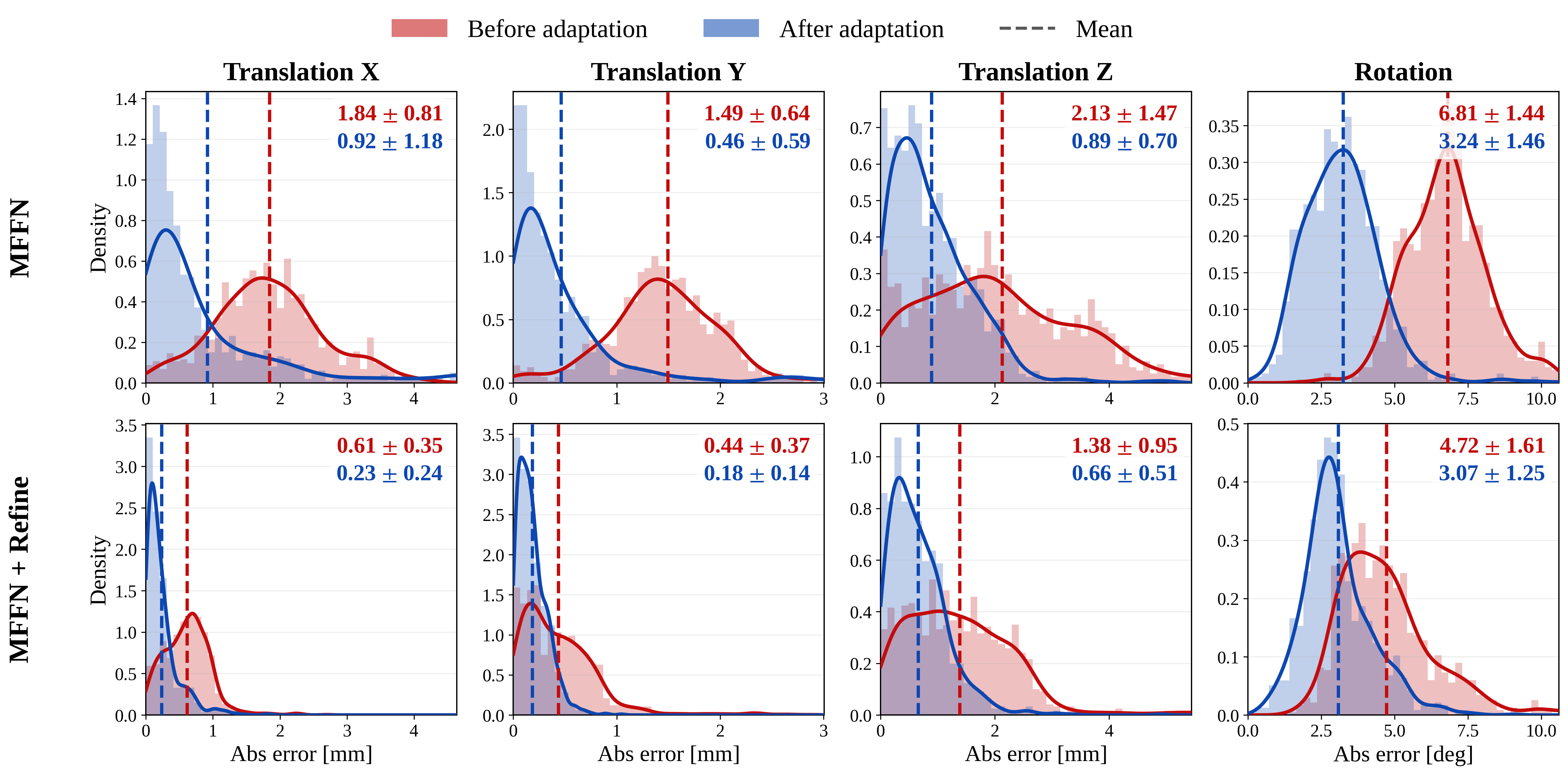}
 \caption{Error distributions before (red) and after (blue) adaptation over the $1{,}000$ real validation samples. Rows show MFFN and MFFN + Refinement. The first three columns show the axis-wise TCP translation errors, and the last column the axis--angle rotation error. Solid curves are kernel density estimates of each distribution, dashed lines mark the means, and each panel is annotated with its mean and standard deviation. Adaptation reduces the mean error across every translation axis and in rotation.}
 \label{fig:adaptation_result}
\end{figure*}

\paragraph{Pseudo ground-truth refinement and adaptation}
Iterative gradient descent initialized from $\hat{T}^{(init)}$ minimizes
$\mathcal{L}_{align}$ and yields a geometrically aligned pose $\hat{T}^{*}$.
The complete procedure is summarized in \Cref{alg:tool_estimation}.
The resulting poses are stored as pseudo-GT in $\mathcal{D}_{pseudo}$, which
contains 150 unlabeled real stereo pairs collected separately from the
$1{,}000$-sample validation set of \Cref{sec:real_validation}.
\Cref{fig:pseudo_gt} illustrates the resulting geometric alignment on real
stereo pairs.

For adaptation, the MFFN perception branches are frozen, and only the pose
regression head and refinement module are fine-tuned using
$\mathcal{D}_{pseudo}$. Freezing the perception branches preserves the
geometric features used to construct the pseudo-GT, while fine-tuning the
pose-regression components adapts their mapping to the real-domain geometry.
The procedure requires no manual pose annotations. Its effect is evaluated
directly on the held-out real validation set in
\Cref{sec:real_validation}.

 \begin{table}[t!]
\centering
\caption{Pose estimation performance with and without self-supervised adaptation on 1{,}000 real samples measured on the deployed TensorRT FP16 pipeline.}
\label{tab:adaption_results}
\setlength{\tabcolsep}{5pt}
\renewcommand{\arraystretch}{1.0}
\resizebox{\columnwidth}{!}{%
\begin{tabular}{lcc}
\hline
Method
& Trans. Error [mm]
& Rot. Error [deg] \\
\hline
MFFN (w/o adapt)
& 3.47 $\pm$ 1.17
& 6.81 $\pm$ 1.44 \\
MFFN + Refine (w/o adapt)
& 1.70 $\pm$ 0.87
& 4.72 $\pm$ 1.61 \\ 
\hline
MFFN (Adapt)
& 1.57 $\pm$ 1.27 
& 3.24 $\pm$ 1.46 \\
MFFN + Refine (Adapt)
& \textbf{0.78 $\pm$ 0.50}
& \textbf{3.07 $\pm$ 1.25} \\
\hline
\end{tabular}%
}
\end{table}

\subsection{Effect of Jaw Case Attachment on Pose-Estimation Accuracy}
\label{sec:jaw_case}
 The estimator uses no markers; therefore, measuring its accuracy requires an external reference. For validation, detachable jaw cases carrying ChArUco markers are attached (\Cref{fig:real_setup}); the concern is whether this attachment alters the estimation being measured.

The model excludes the marker region from all masks and keypoints (\Cref{fig:dataset_kpts_mask})  and \Cref{tab:jaw_case_attachment} quantifies the remaining influence in the simulation. Including the case geometry  changed the error from only $0.14$~mm\,/\,$0.44^\circ$ to $0.17$~mm\,/\,$0.53^\circ$; therefore, the visual effect of the attachment on the estimate is small in the simulation.

\begin{figure*}[t!]
 \centering
 \includegraphics[width=0.72\linewidth]{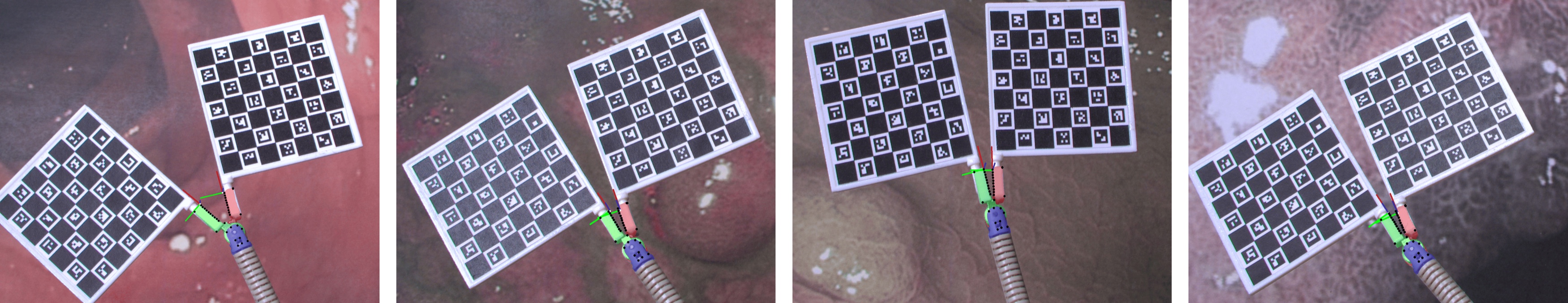}
\caption{Real-world validation. Predicted masks, keypoints, and 6-DoF poses (coordinate axes) overlaid on real stereo images, showing accurate alignment between the predicted geometry and real observations under diverse viewpoints and backgrounds. The jaw-mounted ChArUco case provides ground-truth poses for evaluation and is excluded from the predicted masks and keypoints.}
 \label{fig:real_mask_pose}
\end{figure*}

 \begin{table*}[t!]
\centering
\caption{
\textbf{Quantitative comparison of real-world TCP pose estimation performance with representative prior methods.}
Translation and rotation errors are reported together with processing time, validation-set size, visual feature modality, platform, and pose annotation strategy. Our entries report the deployed TensorRT FP16 pipeline.
}
\label{tab:pose_comparison_real}
\setlength{\tabcolsep}{5pt}
\renewcommand{\arraystretch}{1.0}
\footnotesize
\begin{tabular}{p{3cm}c c c c c c c}
\hline
\textbf{Method} 
& \textbf{Trans. Err [mm]} 
& \textbf{Rot. Err [deg]} 
& \textbf{Proc. Time [ms]} 
& \textbf{Val. Size} 
& \textbf{Modality} 
& \textbf{Platform} 
& \textbf{Validation GT} \\ \hline
Zhou et al. (base) \cite{zhou2024markerless}
& $1.63 \pm 0.78$
& $5.02 \pm 1.97$
& $\sim$29$^\dagger$
& 162
& Single (mask)
& Continuum
& Marker \\
Zhou et al. (refine) \cite{zhou2024markerless}
& $1.24 \pm 0.85$
& $3.20 \pm 1.45$
& $\sim$830$^\dagger$
& 162
& Single (mask)
& Continuum
& Marker \\
Ye et al. \cite{Ye_2016}
& $3.14 \pm 1.96$ 
& $6.88 \pm 4.58$ 
& 34.5 
& 1000 
& Single (kpts) 
& dVRK 
& Manual \\
Allan et al. \cite{8295119}
& $3.85 \pm 3.64$ 
& $24.64 \pm 14.90$ 
& 3k--6k 
& 1000 
& Single (mask) 
& dVRK 
& Manual \\ \hline
MFFN (Adapt)
& $1.57 \pm 1.27$ 
& $3.24 \pm 1.46$ 
& 39
& 1000
& Multi 
& Continuum 
& Marker \\
\textbf{MFFN + Refine (Adapt)}
& \textbf{\boldmath $0.78 \pm 0.50$} 
& \textbf{\boldmath $3.07 \pm 1.25$} 
& 52.4
& 1000 
& Multi 
& Continuum 
& Marker \\
\hline
\multicolumn{8}{l}{\footnotesize $^\dagger$Approximate follow-up-frame processing time calculated from the component-wise runtimes reported by Zhou \textit{et al.}: 17\,ms for segmentation,}\\
\multicolumn{8}{l}{\footnotesize 9\,ms for partial-pose estimation, 3\,ms for the kinematics module, and approximately 800\,ms for iterative refinement when applicable.}\\
\multicolumn{8}{l}{\footnotesize The kinematics module requires 20\,ms on the first frame.}\\
\end{tabular}
\end{table*}

\subsection{Time Efficiency and TensorRT Deployment}
\label{sec:time_efficiency}
All runtime measurements were obtained on a single NVIDIA RTX A5000 GPU. In native PyTorch, the complete framework requires $187.0$\,ms of total processing time per stereo pair, corresponding to approximately $5.3$\,Hz (\Cref{tab:time_param}).  We compiled the object detector, MFFN, and  refinement module using TensorRT (FP16) to optimize inference speed. The MFFN  inference time decreases from $131.5$~ms to $31.4$~ms, and detector latency decreases from $33.9$~ms to $6.8$~ms.  Total processing time falls to $52.4$~ms per stereo pair, a $3.6\times$ speedup that  increases the pose update rate to  approximately $19$~Hz.

Over the same $1{,}000$ real frames, the compiled pipeline  achieves an error of $0.78$~mm and $3.07^\circ$, compared to the $0.77$~mm and $3.16^\circ$ in native PyTorch. All real-world and closed-loop experiments (\Cref{sec:real_validation,sec:closed_loop})  were conducted using the optimized pipeline.

\section{Model Validation in Real Environments}
\label{sec:real_validation}
For real-world validation, ChArUco markers attached to each jaw provided the ground-truth poses of Jaws~1 and~2. For visual diversity, four endoscopic surgical images from~\cite{surgical_open_dataset} were printed and placed behind the workspace as backgrounds. For each background condition, a different random motion sequence comprising 250 joint configurations was recorded, yielding 1{,}000 real validation samples in total. We evaluated MFFN and MFFN + Refinement performance before and after the proposed self-supervised adaptation.

\subsection{Performance with and without Self-Supervised Adaptation}
\Cref{tab:adaption_results} summarizes the 1{,}000 real samples.  Self-supervised adaptation reduced the MFFN error from $3.47$~mm\,/\,$6.81^\circ$ to $1.57$~mm\,/\,$3.24^\circ$ and the full framework from $1.70$~mm\,/\,$4.72^\circ$ to $0.78$~mm\,/\,$3.07^\circ$ ($54\%$ translation, $35\%$ rotation).  The two effects were distinct given that the evaluation covered every combination of refinement and adaptation. Adaptation improved translation by $2.2\times$ in both configurations, providing the largest rotation gain when refinement  was absent ($6.81^\circ\!\to\!3.24^\circ$). In contrast, refinement halved the translation in both adaptation states ($1.57\!\to\!0.78$~mm adapted); however, once adapted,  the rotation changed only marginally ($3.24^\circ\!\to\!3.07^\circ$) at an additional cost of  approximately $14$  ms for rendering and refinement (\Cref{tab:time_param}).  In real data, rotation accuracy was influenced primarily by MFFN and adaptation,  whereas refinement contributed mostly to translation. A per-feature analysis  should be conducted in future work. \Cref{fig:adaptation_result} shows that adaptation substantially reduces the systematic offset observed before adaptation across the translation axes and rotation. The result demonstrates the net effect of adaptation on the held-out real validation set without attributing the improvement to a single source of sim-to-real error.

\subsection{Performance Comparison with Previous Methods}
\Cref{tab:pose_comparison_real} compares the adapted model with prior methods~\cite{zhou2024markerless, Ye_2016, 8295119}, and \Cref{fig:real_mask_pose} shows qualitative real-world results.

For comparison, Zhou \textit{et al.}~\cite{zhou2024markerless} reported 1.63\,mm\,/\,$5.02^\circ$ before refinement and 1.24\,mm\,/\,$3.20^\circ$ after iterative render-and-compare refinement, with the refinement stage alone requiring approximately 800\,ms per frame. The adapted stereo MFFN achieved 1.57\,mm\,/\,$3.24^\circ$, and the proposed single-pass refinement improved this to 0.78\,mm\,/\,$3.07^\circ$ at 52.4\,ms per stereo pair. Relative to the iteratively refined result reported by Zhou \textit{et al.}, this corresponds to a 37\% lower translation error at more than $15\times$ lower latency. However, this cross-study comparison is approximate because the platforms, ground-truth procedures, and evaluation scales differ: Zhou \textit{et al.} evaluated 162 configurations, whereas we evaluate 1{,}000 samples under four background conditions. Therefore, the within-platform ablation in \Cref{tab:adaption_results} provides the primary evidence for the improvement, while \Cref{tab:pose_comparison_real} provides cross-study context.

\subsection{Single Pass versus Iterative Refinement}
\label{sec:refine_analysis}
The proposed \emph{single-pass refinement} corrects the MFFN pose with one forward pass of the learned module (\Cref{sec:refine_module}), whereas conventional \emph{iterative render-and-compare}~\cite{zhou2024markerless, liang2025differentiable_surg_pose} refines an initial pose by repeatedly minimizing a rendering-based alignment objective. We compare the two refinement strategies on the same $1{,}000$ real validation samples used in \Cref{sec:real_validation}. Both methods are initialized from the same adapted MFFN pose and use the same predicted masks and keypoints and the same CAD renderer. The iterative baseline minimizes $\mathcal{L}_{align}$ of \Cref{sec:adaptation} using Adam (lr $3\times10^{-3}$) for $1{,}500$ steps per sample in FP32, whereas the single-pass result is obtained from the deployed TensorRT FP16 pipeline.

\begin{figure}[!tb]
 \centering
 \includegraphics[width=0.8\linewidth]{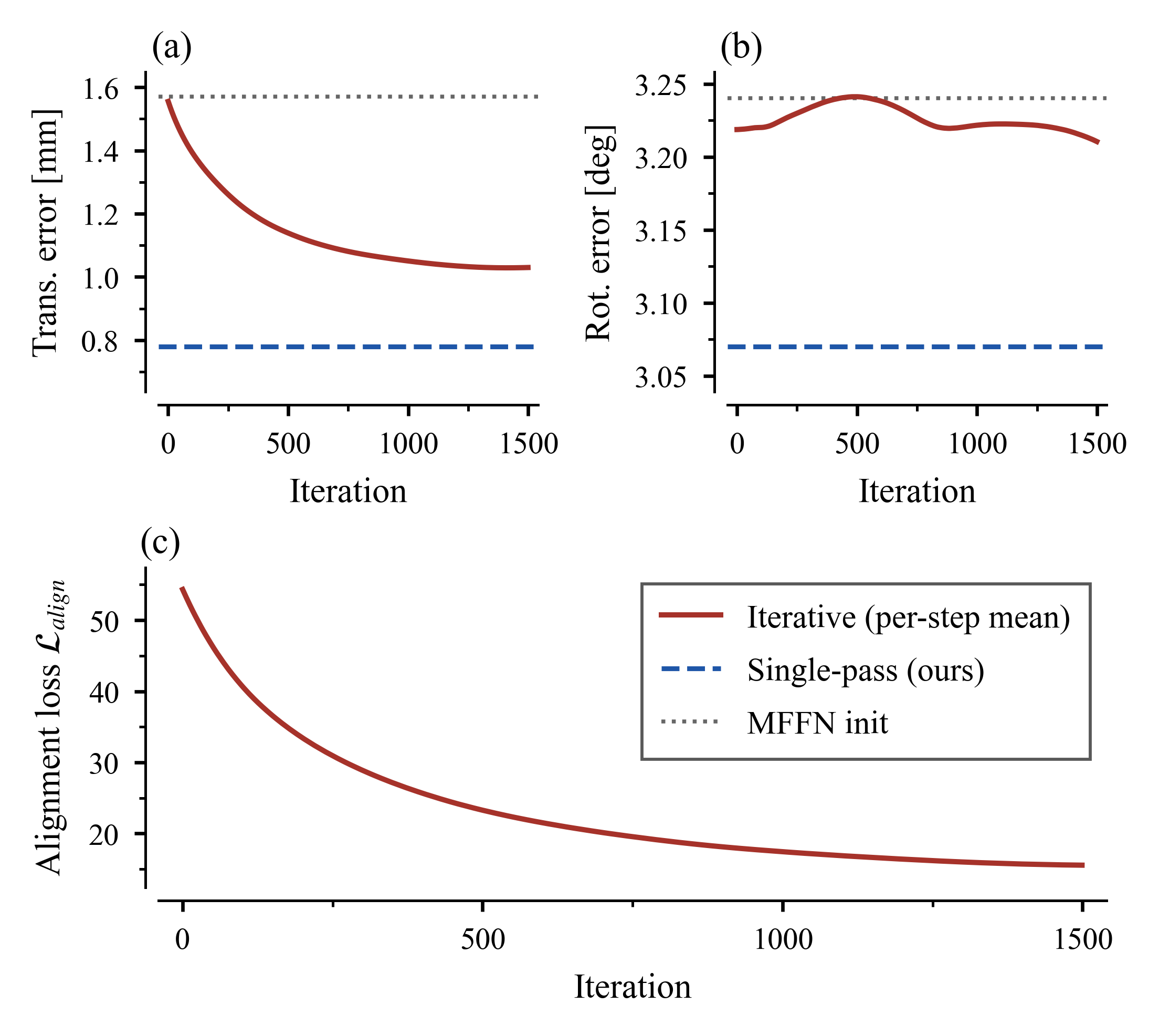}
 \caption{Comparison of the proposed single-pass refinement and conventional iterative render-and-compare on the $1{,}000$ real validation samples. (a) Mean translation error, (b) mean rotation error, and (c) mean alignment loss $\mathcal{L}_{align}$ during iterative optimization. In (a) and (b), the blue dashed and gray dotted lines indicate the proposed single-pass result and the initial adapted MFFN estimate, respectively, while the red solid curve shows the conventional iterative render-and-compare trajectory. Iterative optimization progressively reduces the translation error but remains above the proposed single-pass result, while the rotation error does not decrease monotonically despite the continued reduction of the alignment loss.}
 \label{fig:tto}
\end{figure}

As shown in \Cref{fig:tto}, iterative optimization reduces the mean translation error from the initial $1.57$\,mm to approximately $1.03$\,mm after $1{,}500$ steps. Selecting, for each sample, the iteration with the lowest alignment loss yields $1.01$\,mm\,/\,$3.29^\circ$. The rotation error remains around $3.2^\circ$ and does not decrease monotonically. In contrast, the learned single-pass refinement reduces the same initial pose error to $0.78$\,mm\,/\,$3.07^\circ$. The complete deployed MarUco pipeline, including detection, MFFN inference, rendering, and learned refinement, requires $52.4$\,ms per stereo pair, whereas the iterative refinement alone requires approximately $150$\,s per sample for the $1{,}500$ optimization steps ($\sim$102\,ms per step).

Although $\mathcal{L}_{align}$ continues to decrease throughout iterative optimization, the translation error approaches a plateau and the rotation error does not exhibit a corresponding monotonic decrease (\Cref{fig:tto}). This discrepancy indicates that further optimization of the image-space alignment objective does not necessarily translate into improved 6D pose accuracy. The learned refinement module instead directly regresses a pose correction from the rendered and predicted geometric features. Under the initialization and optimization settings evaluated here, it achieves lower pose error with substantially lower processing time than the iterative baseline.

\section{Closed-Loop Control of the Continuum Manipulator}
\label{sec:closed_loop}
We compared three control baselines: uncompensated open-loop control, PBVS with the ChArUco marker pose as feedback, and PBVS with the MarUco estimate as feedback. Open-loop control converts  reference waypoints into joint commands through  inverse kinematics (\Cref{sec:ik}) and applies them without  feedback. Marker-based PBVS serves as the  reference sensor baseline. MarUco-based PBVS runs in the same loop,  relying exclusively on the markerless pose from the deployed pipeline of \Cref{sec:time_efficiency}.

\subsection{Jacobian-Based Position-Based Visual Servoing}
\label{sec:servoing_law}
 At each control cycle $k$, the observed TCP pose, supplied either by the markerless estimator or  the ChArUco marker tracker, is transformed into the robot base frame through  calibrated extrinsics: ${}^{base}\mathbf{T}^{obs}_{tcp,k} = {}^{base}\mathbf{T}_{cam}\,{}^{cam}\mathbf{T}^{obs}_{tcp,k}$.

The joint configuration is computed as $\mathbf{q}_k = \mathrm{IK}({}^{base}\mathbf{T}^{obs}_{tcp,k})$, and the geometric Jacobian $\mathbf{J}(\mathbf{q}_k)$ is evaluated at this configuration.

The error in the desired pose ${}^{base}\mathbf{T}^{*}_{tcp}$
 \begin{equation}
\mathbf{T}_{err,k}
=
({}^{base}\mathbf{T}^{obs}_{tcp,k})^{-1}
\, {}^{base}\mathbf{T}^{*}_{tcp},
\end{equation}
is expressed in the base frame as a stacked  task space error 
\begin{equation}
\mathbf{e}_k =
\begin{bmatrix}
{}^{base}\mathbf{R}_{tcp,k}\,\mathbf{p}_{err,k} \\
{}^{base}\mathbf{R}_{tcp,k}\,\mathrm{rotvec}(\mathbf{R}_{err,k})
\end{bmatrix},
\end{equation}
where $\mathbf{p}_{err,k}$ and $\mathbf{R}_{err,k}$  represent the translational and rotational parts of $\mathbf{T}_{err,k}$, and $\mathrm{rotvec}(\cdot)$ maps  the rotation matrix into its axis-angle representation.
The joint command is updated  using the Jacobian pseudoinverse
 \begin{equation}
\mathbf{q}_{k+1} = \mathbf{q}_k + \mathbf{J}(\mathbf{q}_k)^{\dagger}\,\alpha\,\mathbf{e}_k,
\end{equation}
with a gain $\alpha = 0.3$, selected for stable convergence under the kinematic model mismatch induced by hysteresis.

To compare the effect of the pose-feedback source, both the marker-based and MarUco-based conditions use the same conventional fixed-gain PBVS control law and the same gain. The marker-based condition provides externally measured 6D pose feedback and serves as the reference PBVS baseline, whereas the MarUco condition replaces this measurement with the proposed markerless 6D pose estimate.

The loop iterates until the error norms cross a convergence threshold or the 100-iteration cap. Marker-based servoing uses thresholds of $1.3$\,mm and $0.03$\,rad. Because MarUco uses estimated pose feedback, the stopping thresholds were relaxed relative to the marker-based baseline to account for pose-estimation error. The thresholds were initially set using the mean estimation error observed during preliminary validation and were subsequently verified in servoing trials. The resulting thresholds were $2.1$\,mm and $0.0782$\,rad ($4.48^\circ$). These thresholds are used solely as stopping criteria. Terminal errors for all control conditions are evaluated using the same external marker reference.

 \begin{figure*}[t!]
 \centering
 \includegraphics[width=0.72\linewidth]{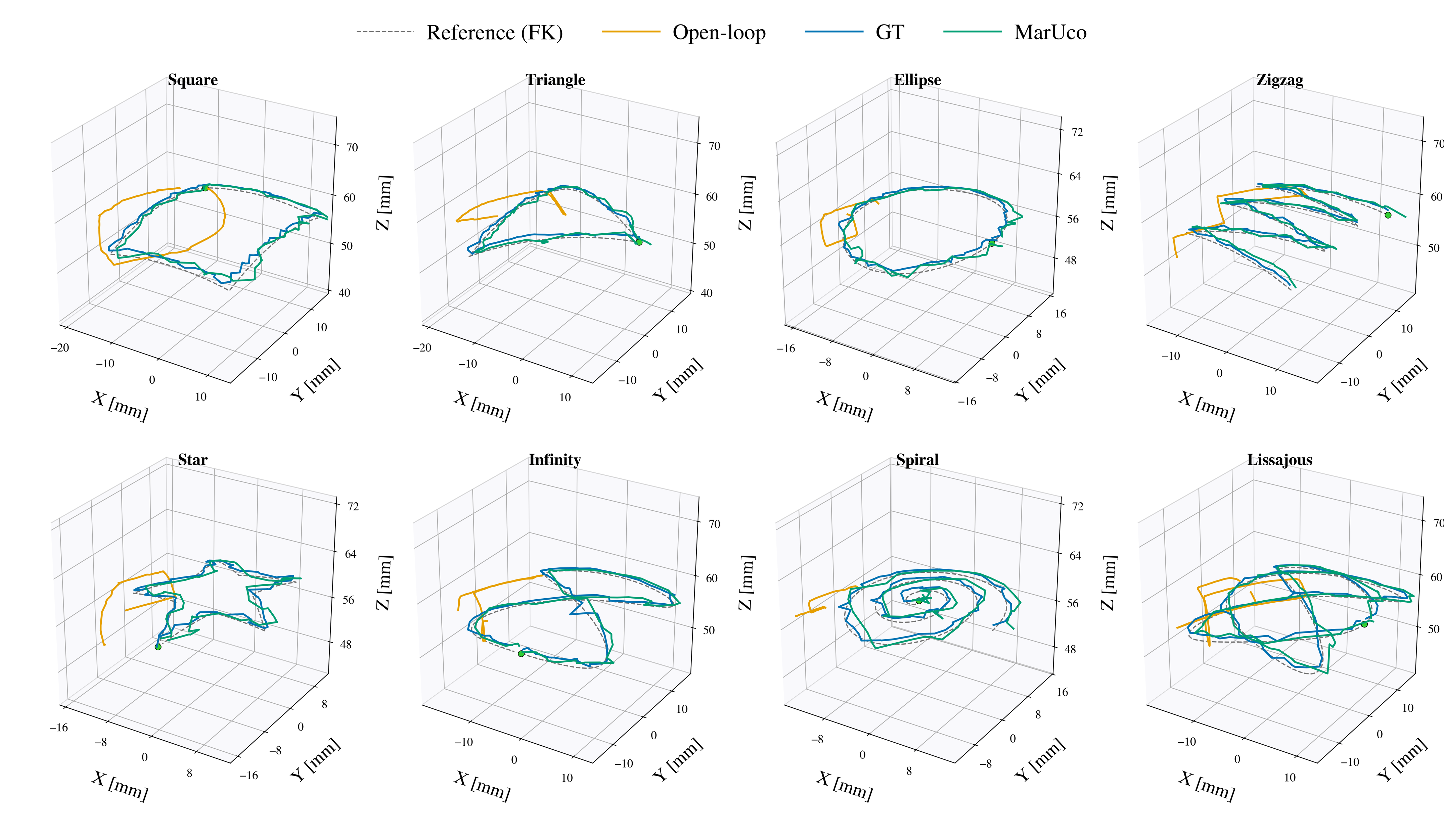}
\caption{Three-dimensional TCP motion along the eight reference paths. The reference path from forward kinematics is a dashed grey, open-loop control (Open-loop) orange, marker-based servoing (GT) blue, and MarUco-based servoing (MarUco) green. MarUco-based servoing follows the marker-based baseline closely across all eight paths, whereas the open-loop control accumulates a large, uncorrected drift.}
 \label{fig:visual_trajectory}
\end{figure*}

\begin{figure*}[t!]
 \centering
 \includegraphics[width=0.8\linewidth]{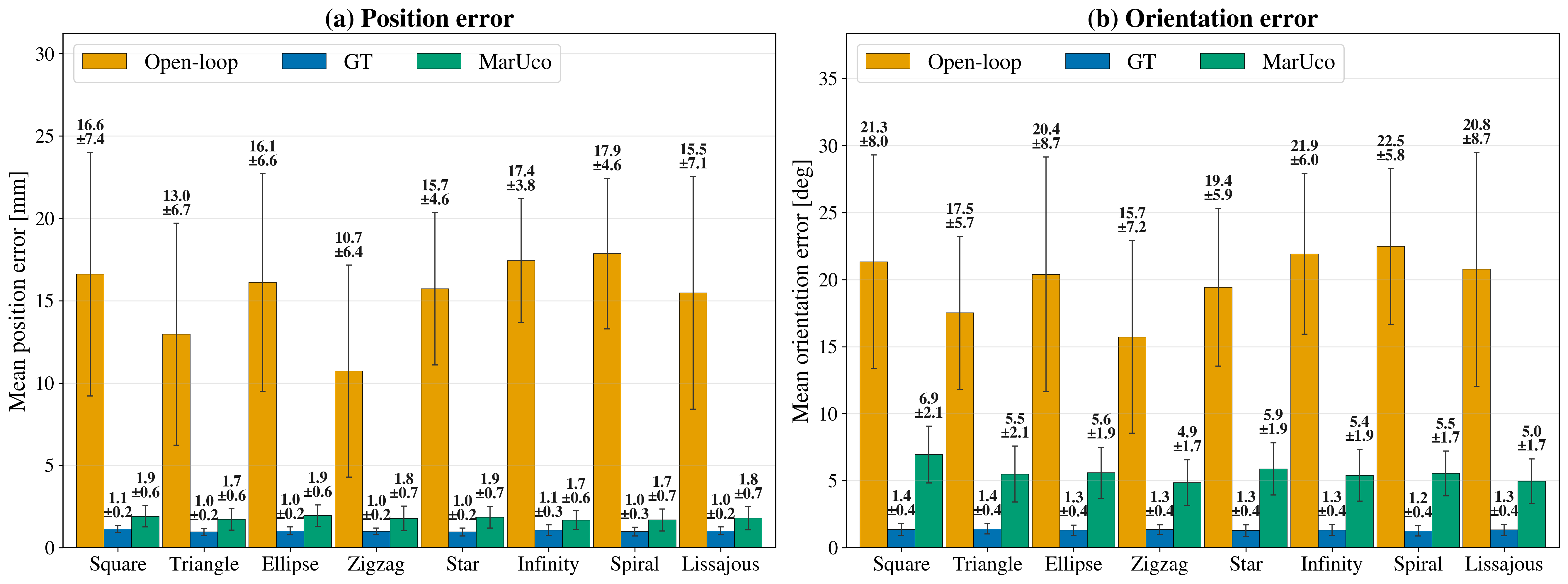}
\caption{Closed-loop path-following accuracy across the eight reference paths. (a) Mean position error and (b) mean orientation error for open-loop control (Open-loop), marker-based servoing (GT, the reference-sensor baseline), and servoing driven by the proposed markerless estimate (MarUco). Bars are means and error bars denote one standard deviation over the waypoints of each path.}
 \label{fig:servoing_bars}
\end{figure*}

\subsection{Path Following Tasks}
\label{sec:traj_tracking}
 We evaluated path following on eight reference paths with  square, triangle, ellipse, zigzag, star, infinity, spiral, and Lissajous curve geometries, each consisting of 120 waypoints. These paths span diverse geometries, from straight segments with sharp reversals to smooth and compound curves. 

For each waypoint, the servoing loop runs until its stopping criterion is  satisfied. The reported error is the marker-measured TCP error at termination  averaged over all waypoints.  Waypoints that hit the iteration cap before convergence contribute their residual errors to the average, so the mean error can exceed the stopping threshold. The quantitative results are shown in \Cref{fig:servoing_bars}.

Averaged across the eight paths, MarUco-based servoing reaches a mean translation error of $1.8$\,mm and a mean rotation error of $5.6^\circ$  against $15.4$\,mm and $19.9^\circ$ for open-loop control,  thereby corresponding to a $88\%$ and $72\%$ reduction, respectively. The per-path translation remained within $1.7$--$1.9$  mm across eight paths, whereas the open-loop translation  varied from $10.7$ to $17.9$  mm. Marker-based servoing, the  reference sensor baseline, reached $1.0$\,mm and $1.3^\circ$, respectively, under a tighter stopping threshold. \Cref{fig:visual_trajectory} shows MarUco-based servoing following the marker-based baseline closely across all paths.

Marker-based servoing retains a mean terminal rotation error of $1.3^\circ$ under the same fixed-gain PBVS controller and its stopping rule. This result indicates that terminal accuracy is influenced by the controller and stopping criterion in addition to the pose-feedback source. The higher terminal rotation error of MarUco-based servoing reflects the combined effect of replacing marker-based feedback with the estimated pose and using the relaxed stopping threshold defined for the MarUco-based loop.

\section{Discussion}
\label{sec:discussion}

\subsection{Generalization to Other Continuum and Compliant Robots}

Structurally, adapting MarUco to another continuum or compliant robot would
require replacing three robot-specific components: (i) a robot model comprising
the pseudo-rigid-body URDF and corresponding link meshes, (ii) distal keypoint
definitions, and (iii) the extrinsic calibration
(\Cref{sec:fk,sec:gt_annotation,sec:calibration}). The robot model only needs
to provide kinematically valid configurations for rendering because the
estimator learns the mapping from camera observations directly to end-effector
pose, rather than the actuation-to-configuration mapping affected by hysteresis
and friction (\Cref{sec:sim_setup}).

Given these components, the same data-generation, training, and self-supervised
adaptation procedure could, in principle, be applied to another platform
without frame-wise manual pose annotation. Transfer to a new platform would
still require regenerating synthetic data using the new robot model, retraining
the pose estimator, and repeating extrinsic calibration and real-domain
adaptation. The transferability discussed here follows from the structure of
the framework rather than cross-platform experimental evidence. Systematic
validation across multiple continuum and compliant robots remains future work.

\subsection{Scope and Future Directions}

The scope of this study is a markerless 6D pose-estimation framework and its
integration with conventional position-based visual servoing on a physical
continuum manipulator.

Robustness in realistic surgical scenes is beyond the scope of this study.
Factors such as smoke, blood, tissue interaction, and overlapping instruments
may degrade the masks and keypoints used by the pose estimator. Evaluating and
improving robustness under such conditions will require dedicated
surgical-scene validation and is left for future work.

Finally, we adopted a conventional fixed-gain PBVS controller
($\alpha=0.3$) to evaluate whether the proposed markerless pose estimates can
serve as feedback for closed-loop control. Marker-based reference feedback
still resulted in approximately $1.0$\,mm\,/\,$1.3^\circ$ terminal error under
the selected gain and stopping rule (\Cref{sec:traj_tracking}), indicating
that the final closed-loop accuracy is determined by both perception and
control. Combining MarUco with more advanced controllers and adaptive stopping
criteria may further improve closed-loop performance.

\section{Conclusion}

This paper presented MarUco, a markerless 6D pose estimation framework for
closed-loop control of continuum manipulators using only stereo vision during
operation. A pseudo-rigid-body URDF and Isaac Sim pipeline generated
200{,}000 stereo pairs with automatically annotated segmentation masks,
bounding boxes, keypoints, and 6D poses. A cross-view multifeature fusion
network combined with single-pass rendering-based refinement achieved
$0.14$\,mm\,/\,$0.44^\circ$ on synthetic data.

For deployment on the physical system, kinematics-free extrinsic calibration
and self-supervised adaptation using 150 unlabeled real stereo pairs mitigated
the systematic pose error observed after sim-to-real transfer. Across
1{,}000 real validation samples, the complete framework achieved
$0.78$\,mm translation error and $3.07^\circ$ rotation error at
$52.4$\,ms per stereo pair. From the same initial MFFN pose, the learned
single-pass refinement achieved lower translation error than the evaluated
iterative render-and-compare baseline while avoiding iterative optimization.

When integrated with conventional position-based visual servoing, MarUco
followed eight reference paths with a mean terminal translation error of
$1.8$\,mm and a mean rotation error of $5.6^\circ$, corresponding to
$88\%$ and $72\%$ reductions relative to uncompensated open-loop control.
To the best of our knowledge, this is the first markerless position-based
visual servoing framework demonstrated on a continuum manipulator.

The present study is limited to a single robotic platform and controlled
laboratory conditions. Future work will investigate robustness under more
challenging surgical scenes, validation across different continuum and
compliant robots, and integration with more advanced closed-loop controllers.

\section*{Acknowledgment}
All sections of this article were written by the authors, and generative AI systems assisted the revision as follows. Claude (Anthropic) and GPT (OpenAI) were used for sentence-level rewording and editing for precision, consistency, and clarity throughout the manuscript. Claude (Anthropic) was additionally used to implement the visualization code that renders  \Cref{fig:adaptation_result,fig:tto,fig:visual_trajectory,fig:servoing_bars} from the authors' experimental data. All methods, experimental data, and technical results were produced by the authors, and every AI-assisted output was reviewed and approved by the authors.

\bibliographystyle{IEEEtran}
\bibliography{ref_arxiv}
 \end{document}